\documentclass[runningheads]{llncs}

\usepackage{eccv}

\usepackage{eccvabbrv}

\usepackage{svg}
\usepackage{graphicx}
\usepackage{subcaption}
\usepackage{booktabs}
\usepackage{mathtools}
\usepackage{algorithm}
\usepackage{algpseudocode}
\usepackage[dvipsnames]{xcolor}
\usepackage{tikz}
\usepackage{xcolor}
\usepackage{stix}
\usepackage{varwidth}%
\usepackage{tabularx,booktabs}

\usepackage{multirow}

\definecolor{squarecolor}{HTML}{7EC07F}
\definecolor{circlecolor}{HTML}{FFD280}

\newcolumntype{n}{X}
\newcolumntype{s}{>{\hsize=.6\hsize}X}
\newcolumntype{x}{>{\hsize=1.4\hsize}X}

\usepackage[accsupp]{axessibility}  %

\usepackage{hyperref}

\usepackage{orcidlink}

\usepackage{amsmath}

\DeclareMathOperator*{\argmin}{arg\,min}

\begin{document}

\title{Rejection Sampling IMLE: Designing Priors for Better Few-Shot Image Synthesis} 

\titlerunning{Rejection Sampling IMLE}

\author{
Chirag Vashist\orcidlink{0000-0003-3270-1533} \and
Shichong Peng\orcidlink{0009-0005-8404-6392} \and
Ke Li\orcidlink{0000-0002-3229-271X}
}

\authorrunning{C.~Vashist et al.}

\institute{APEX Lab \\ 
School of Computing Science \\
Simon Fraser University\\
\email{\{\href{mailto:chirag_vashist@sfu.ca}{chirag\_vashist},
\href{mailto:shichong_peng@sfu.ca}{shichong\_peng},
\href{mailto:keli@sfu.ca}{keli}\}@sfu.ca}
}

\maketitle

\begin{abstract}

An emerging area of research aims to learn deep generative models with limited training data. Prior generative models like GANs and diffusion models require a lot of data to perform well, and their performance degrades when they are trained on only a small amount of data. A recent technique called Implicit Maximum Likelihood Estimation (IMLE) has been adapted to the few-shot setting, achieving state-of-the-art performance. However, current IMLE-based approaches encounter challenges due to inadequate correspondence between the latent codes selected for training and those drawn during inference. This results in suboptimal test-time performance. We theoretically show a way to address this issue and propose RS-IMLE, a novel approach that changes the prior distribution used for training. This leads to substantially higher quality image generation compared to existing GAN and IMLE-based methods, as validated by comprehensive experiments conducted on nine few-shot image datasets.

\end{abstract}
\begin{figure}[ht]
    \centering
    \includegraphics[width=\linewidth]{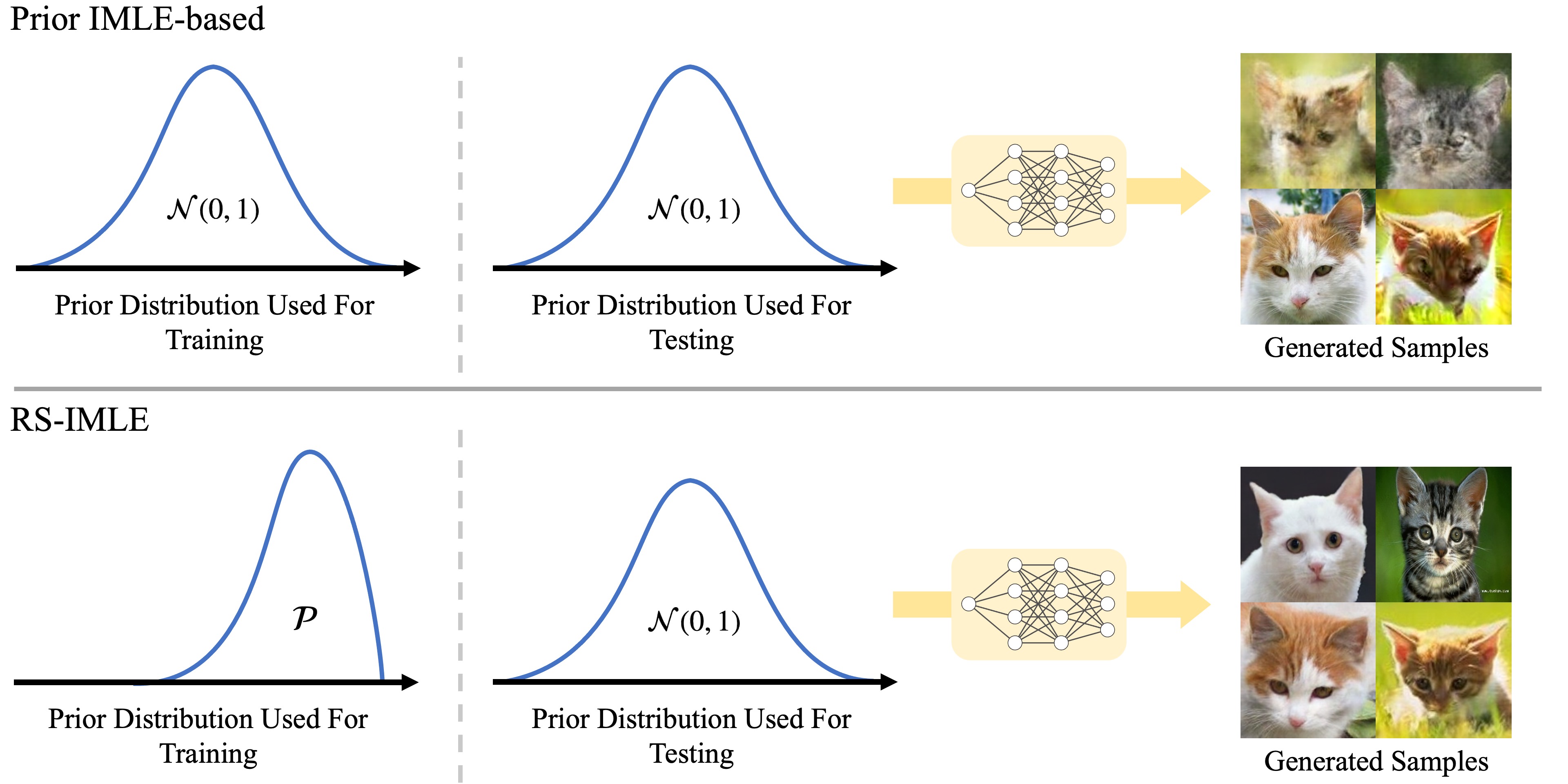}
    \caption{IMLE is an implicit generative model that maps a latent code sampled from a prior distribution to an image output. In previous IMLE-based methods, both the training and testing phases adopt a standard normal distribution as the prior distribution. However, this approach often results in poor generalization during inference. To address this limitation, we introduce RS-IMLE, which uses rejection sampling to alter the prior distribution used for training to a different distribution $\mathcal{P}$. This modification significantly enhances the quality of generated images during testing.}
    \label{fig:teaser}
\end{figure}

\section{Introduction}
\label{sec:intro}

Recent years have witnessed significant advances in image synthesis, driven by the development of a broad variety of powerful generative models. Generative adversarial networks (GANs)~\cite{goodfellow2014generative,biggan,karras2019style,stylegan2,stylegan3}, variational autoencoders (VAEs)~\cite{kingma2013auto,Vahdat2020NVAEAD,vdvae,Razavi2019GeneratingDH}, diffusion models~\cite{Dhariwal2021DiffusionMB,Ho2020DenoisingDP}, score-based models~\cite{Song2021ScoreBasedGM,Song2019GenerativeMB}, normalizing flows~\cite{Dinh2017DensityEU,Kobyzev2021NormalizingFA,Kingma2018GlowGF}, and autoregressive models~\cite{Salimans2017PixelCNNIT,Oord2016PixelRN,Oord2016ConditionalIG,Esser2020TamingTF} have demonstrably improved the quality of synthesized images, often achieving photorealism. 
However, to achieve this high fidelity, generative models often require large amounts of training data. 

In some scenarios, there are not a lot of training examples available.  Suppose we want to emulate the types of edits that a user manually made to a few images. In this scenario, we only have access to a limited number of training examples to begin with. 
In other cases, the training data can be hard to collect. In autonomous driving, synthesizing images for rare conditions like near misses can be challenging. There are also cases where collecting data is expensive. Suppose we want to train a 3D generative model, which requires 3D objects. Creating these 3D objects often involves expensive manual labour or running reconstruction algorithms. In this paper, we aim to tackle the problem of high-quality image synthesis using limited training data.

\begin{figure*}[!t]
  \captionsetup[subfigure]{labelformat=parens}
  \begin{subfigure}[t]{0.45\textwidth}
    \centering
    \includegraphics[width=\linewidth]{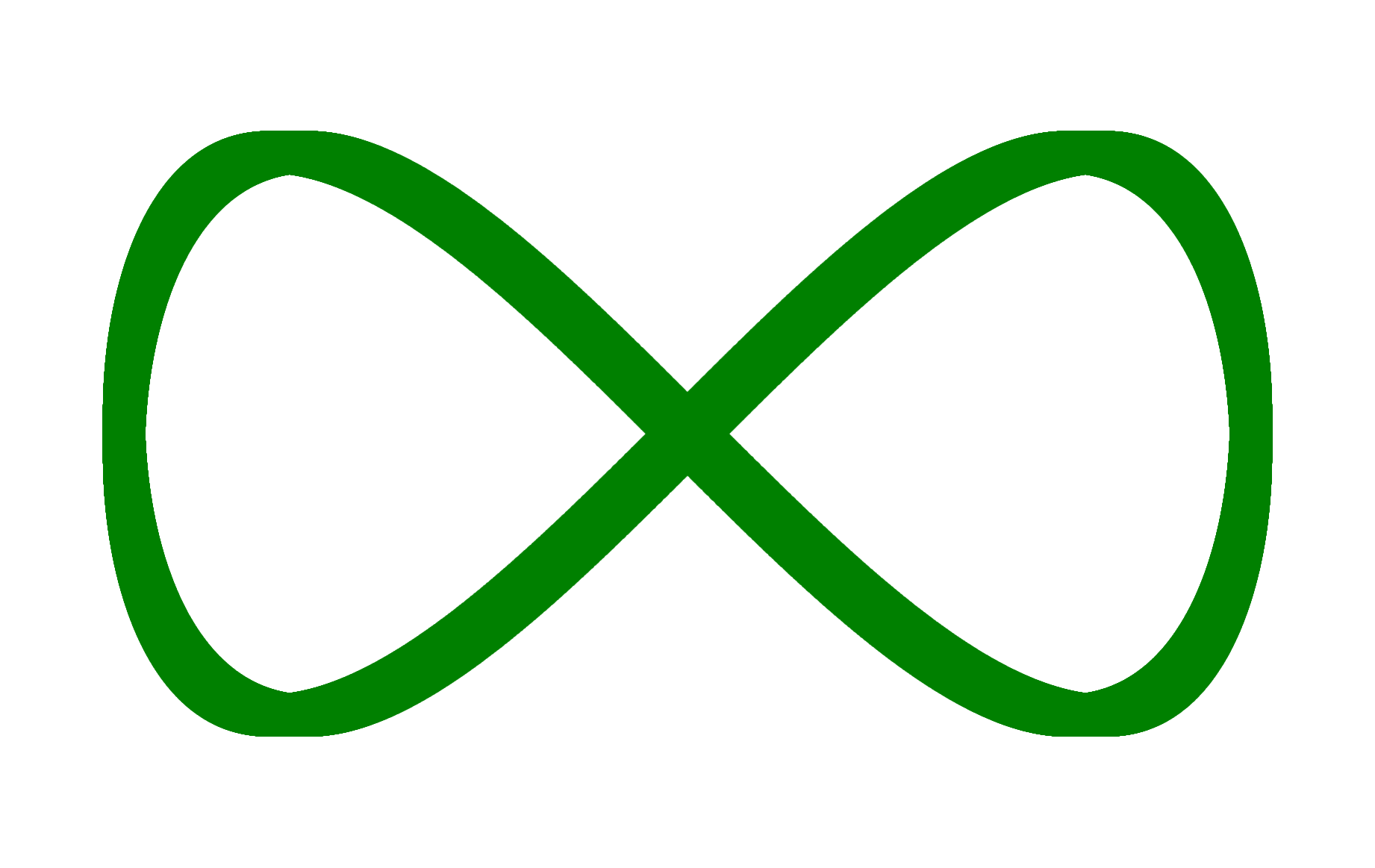}
    \caption{{Dataset containing 10K data points}}
    \label{fig-diffusion:10k}
  \end{subfigure}
  \hfill
  \begin{subfigure}[t]{0.45\textwidth}
    \centering
    \includegraphics[width=\linewidth]{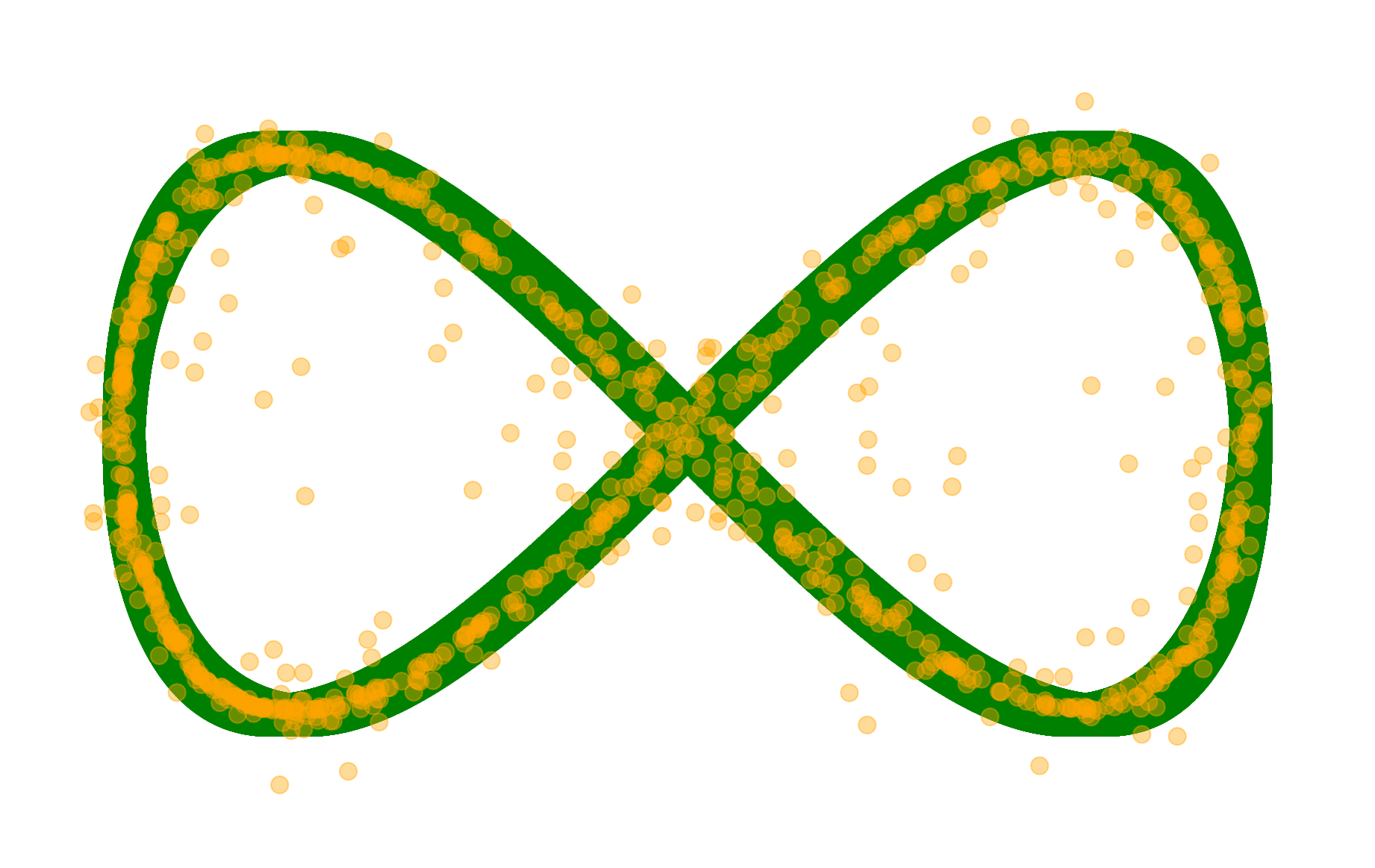}
    \caption{{Samples from diffusion model trained on 10K datapoints}}
    \label{fig-diffusion:10k-trained}
  \end{subfigure}

  \begin{subfigure}[t]{0.45\textwidth}
    \centering
    \includegraphics[width=\linewidth]{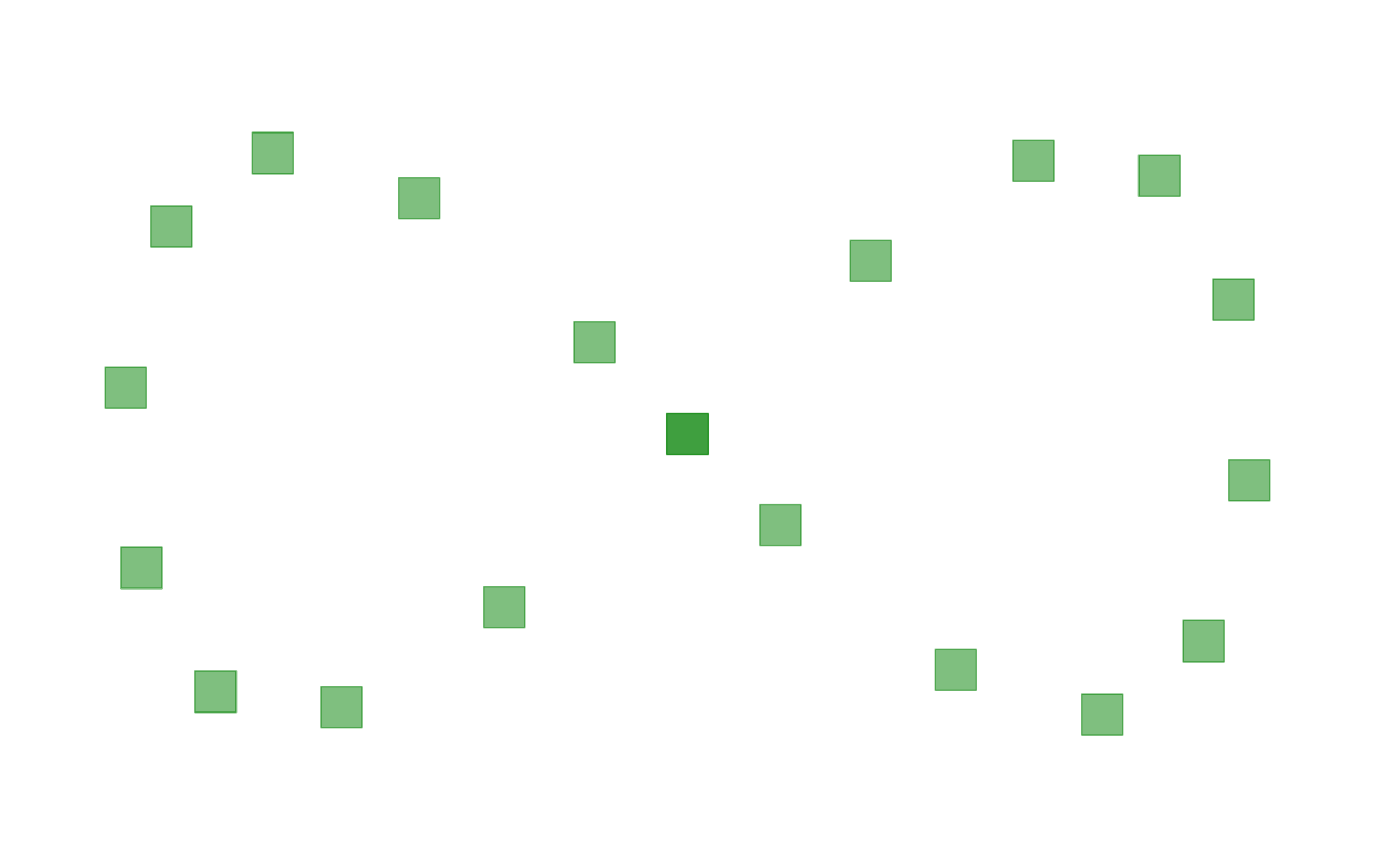}
    \caption{{Dataset containing 20 data points} }
    \label{fig-diffusion:20}
  \end{subfigure}
  \hfill
  \begin{subfigure}[t]{0.45\textwidth}
    \centering
    \includegraphics[width=\linewidth]{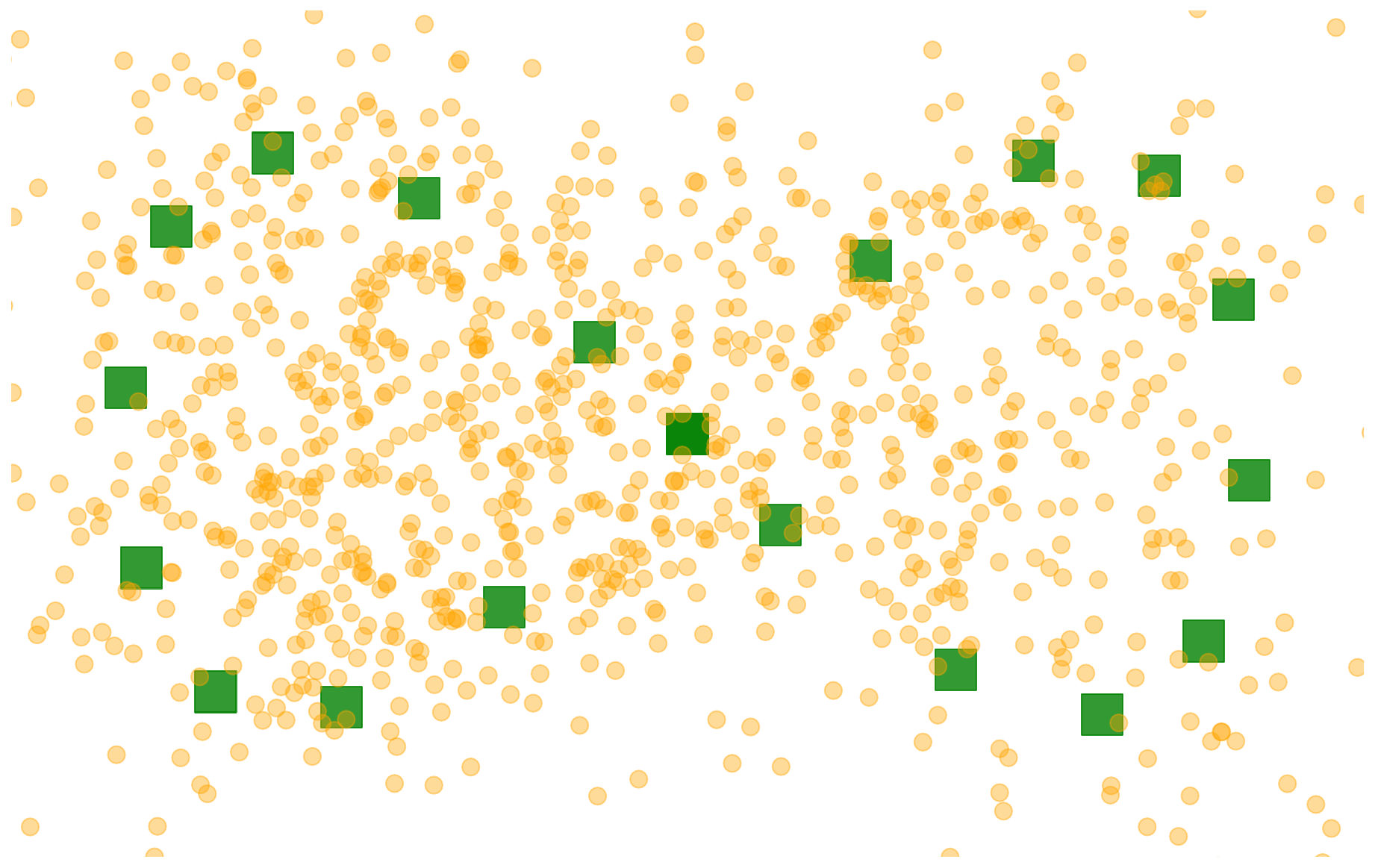}
    \caption{{Samples from diffusion model trained on 20 datapoints}}
    \label{fig-diffusion:20-trained}
  \end{subfigure}

  \caption{Comparison between performance of diffusion models on large-scale and few-shot setting. We have two 2D datasets of the same shape (infinity symbol) but different number of data points: 10K data points \ref{fig-diffusion:10k} and 20 data points \ref{fig-diffusion:20}. We train the \emph{same model} but get very different performance. For the few-shot case (20 data points), the diffusion model fails to learn a distribution that matches the data distribution.
  Data points are denoted by ${\color{ForestGreen}\mdblksquare}$ and samples are denoted by ${\color{circlecolor}\mdblkcircle}$.}
  \label{diffusion-compare}
\end{figure*}

The limited availability of training data in this context makes it crucial for generative models to fully leverage every provided example. Generative models that perform well in the large-scale setting, do not perform well in the few-shot setting. In diffusion models, the marginal likelihood under the forward process is a mixture of isotropic Gaussians. This modeling assumption smooths out the learned manifold along all directions, including those that are orthogonal to the actual data manifold. This becomes particularly problematic when there are a limited number of training examples (Figure \ref{diffusion-compare}). Hence, implicit generative models are commonly employed for few-shot generation, with the generator in GANs ~\cite{DBLP:fastgan,mixdl,fakeclr,yang2022FreGAN,ReGAN} serving as a notable example. However, GAN-based methods continue to be afflicted by mode collapse. Mode collapse occurs when the generator network fails to capture the full training data distribution and instead produces a limited subset of outputs. This phenomenon is especially problematic in scenarios where only a small number of training examples are provided. 

Implicit Maximum Likelihood Estimation (IMLE)~\cite{li2018implicit} is an alternative to the GAN objective and has shown promising results in addressing mode collapse. In contrast to GANs, which aim to make each generated image resemble some training data, IMLE instead ensures that each training image has \emph{some} generated sample close to it, and therefore cannot drop any of the modes present in the training data. Adaptive IMLE~\cite{adaptiveIMLE} further extends IMLE to the few-shot image synthesis setting and achieves state-of-the-art generated image quality and mode coverage. 

\begin{figure*}[!t]
    \centering

  \begin{subfigure}[t]{0.45\textwidth}
    \centering
    \includegraphics[width=\linewidth]{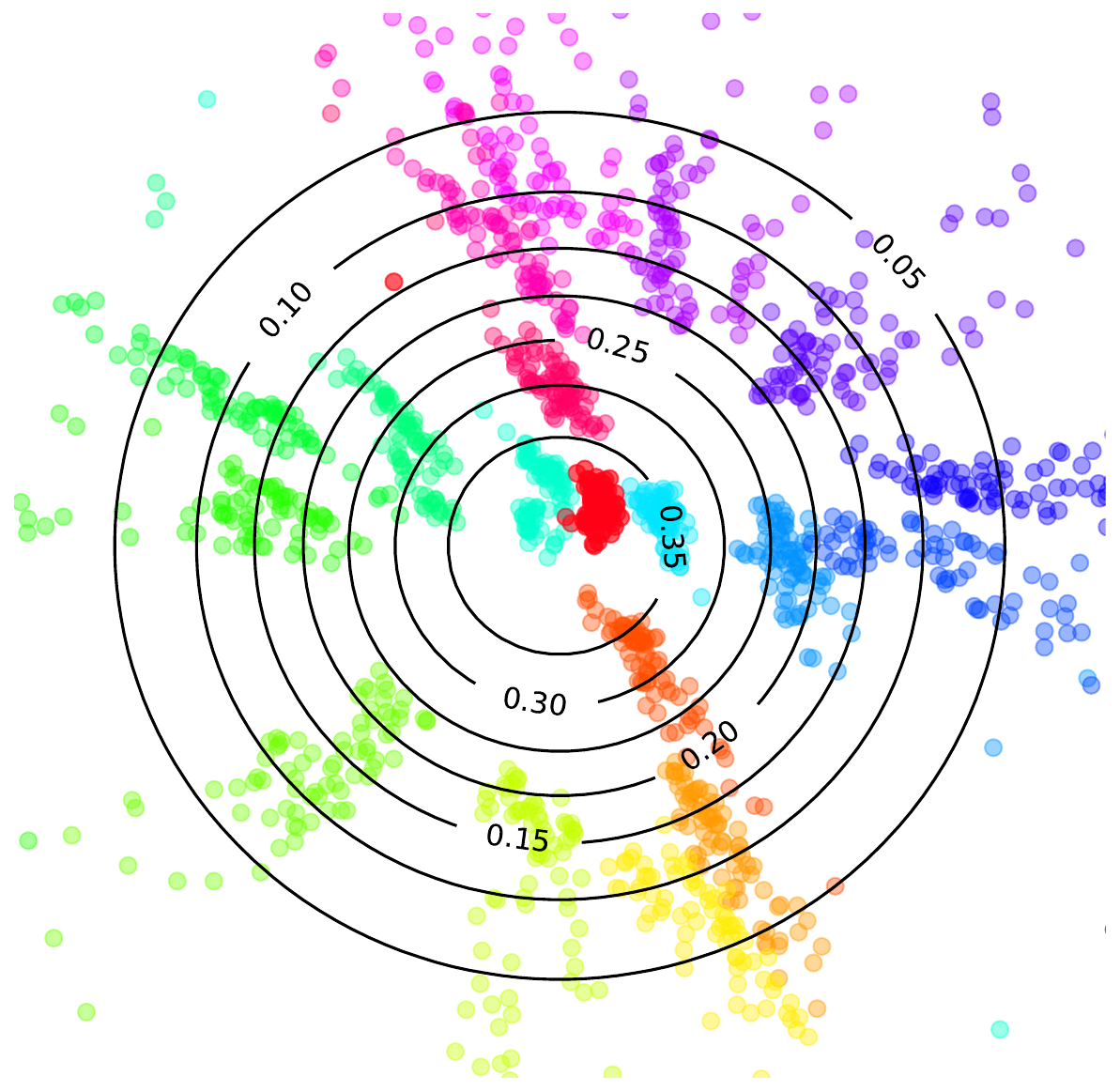}
    \caption{{Latent space of model trained by IMLE objective using standard normal prior}. Dots represent the points selected by the model \emph{over the course of training}, with dots of the same colour belonging to the same data point. The contours of standard normal distribution have been shown for comparison.}
    \label{fig-banner-latent-imle}
  \end{subfigure}
  \hfill
  \begin{subfigure}[t]{0.45\textwidth}
    \centering
    \includegraphics[width=\linewidth]{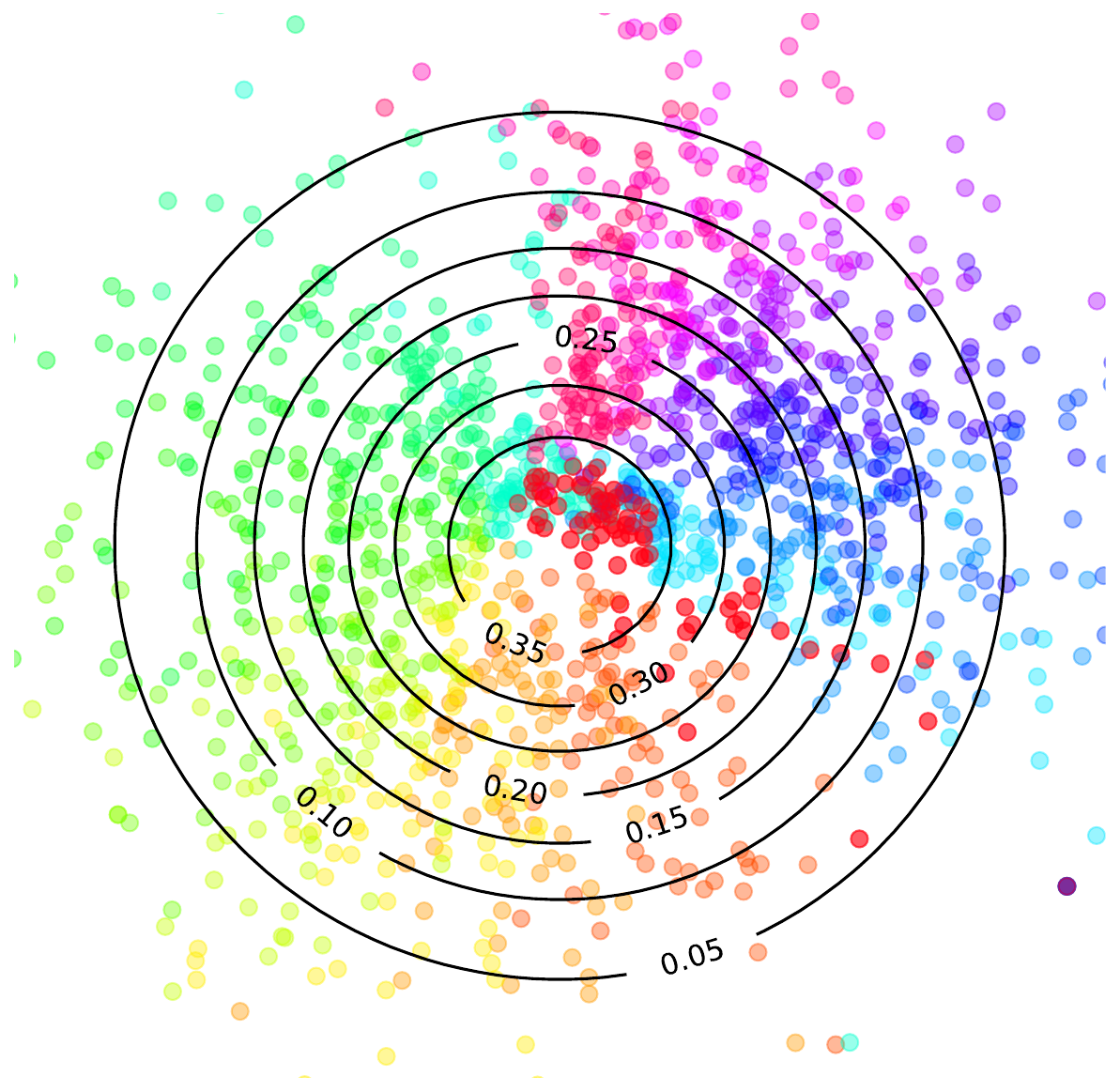}
    \caption{{Latent space of model trained by RS-IMLE objective using prior obtained via rejection sampling}. Compared to latent space of model trained by IMLE, our method over the course of training, samples latent codes that follow the distribution at test time more faithfully.}
    \label{fig-banner-latent-eps-imle}
  \end{subfigure}

  \caption{Difference between the latent codes picked by IMLE and RS-IMLE over the course of training.}
  \label{fig-banner}
\end{figure*}

However, in existing IMLE-based approaches, we observe that the latent codes used during training and those sampled during testing have different distributions, even though the same prior is used during training and testing. This phenomenon arises because of how IMLE selects latent codes during training. Some regions of the latent space are consistently rarely picked for training, despite having a high likelihood under the prior distribution (often the standard Gaussian). This is illustrated in Fig.~\ref{fig-banner-latent-imle}. 
Consequently, at test time, when latent codes drawn from the prior happen to fall in these regions, they yield low-quality samples that are far from the real data points, as illustrated in Fig.~\ref{fig:teaser}. 

This issue has been observed in other generative models like VAEs. Hoffman et. al \cite{ELBO_surgery_2016} show that in practice the prior distribution $p(z)$ and the approximate posterior $q(z)$ are substantially different. Subsequent work \cite{saha2023vae} attempt to mitigate this mismatch by minimizing the KL-divergence between the prior distribution and the aggregate posterior. This approach in turn has its own drawbacks as it can lead to posterior collapse, which dimishes the generative capabilities of VAEs.

Rather than trying to change the objective like in the previous line of work, we address this issue by carefully choosing a different prior so that the samples selected for training have a distribution more similar to those sampled at inference. Our method, which we call Rejection Sampling IMLE or RS-IMLE for short, demonstrably improves coverage of the latent space used during training, thereby ensuring better alignment with the prior as shown in Fig.~\ref{fig-banner-latent-eps-imle}. As a result, our method yields higher quality samples during testing compared to existing GAN and IMLE-based methods. We substantiate this claim through theoretical analysis and extensive experiments conducted on nine few-shot image datasets. We achieve an average of 45.9\% decrease in FID ~\cite{fid} across datasets compared to the best baseline.

\section{Related Work}
Training deep generative models with limited data remains a significant challenge. One approach involves adapting a model pretrained on a large-scale auxiliary dataset from similar domains~\cite{Li2020FewshotIG,Zhao2020OnLP,Mo2020FreezeDA,Ojha2021FewshotIG,Wang2020MineGANEK,Sauer2021ProjectedGC}. However, the availability of such large-scale auxiliary datasets across all domains is not guaranteed. Therefore, another emerging line of work focuses on training models from scratch. In this context, due to the scarcity of training data, diffusion models struggle to achieve high-quality generated images and have been demonstrated to be ineffective~\cite{adaptiveIMLE}. As a result, previous works in this area predominantly build on Generative Adversarial Networks (GANs) and design various methods to address the well-known mode collapse issue. Techniques such as ADA~\cite{Karras2019AnalyzingAI} and DiffAug~\cite{Zhao2020DifferentiableAF} aim to expand training data using adaptive and differentiable augmentation strategies. FastGAN~\cite{DBLP:fastgan} introduced a skip-layer excitation module for accelerated training and used self-supervision in the discriminator to enhance feature learning, thereby improving mode coverage of the generator. FakeCLR~\cite{fakeclr} enhances image synthesis by extensive data augmentation and applies contrastive learning solely on perturbed fake samples. FreGAN~\cite{yang2022FreGAN} introduces a frequency-aware model with a self-supervised constraint to avoid generating arbitrary frequency signals. ReGAN~\cite{ReGAN} dynamically adjusts GANs' architecture during training to explore diverse sub-network structures at different training times. However, despite these advances, some degree of mode collapse persists.

In contrast, Implicit Maximum Likelihood Estimation (IMLE)~\cite{li2018implicit} shows promising results in addressing mode collapse through the use of an alternative objective function compared to GANs. Building upon IMLE, Adaptive IMLE~\cite{adaptiveIMLE} adapts this approach to the few-shot image synthesis scenario by introducing individual target thresholds for each training data point. This dynamic adjustment of training progress accounts for varying difficulties across different data points, thereby effectively leveraging the limited training data. In this work, we introduce a novel algorithm, orthogonal to Adaptive IMLE, for sample selection during training.

\section{Method}

\subsection{Background}

In the context of unconditional image synthesis, the primary objective is to learn the unconditional probability distribution of images $p(\mathbf{x})$. This distribution enables the generation of novel synthesized images through sampling. Generator in GANs are represented by a function $T_\theta : Z \rightarrow X$, implemented as a neural network with parameters denoted as $\theta$. The function $T_\theta$ learns a transformation from the latent space $Z$ to the image space $X$ by using adversarial training, which employs a discriminator that tries to distinguish between generated images $T_\theta(\mathbf{z})$ and real images $\mathbf{x}$, while the generator tries to produce increasingly realistic images to deceive the discriminator. However, this objective often leads to mode collapse, a well-known issue of GANs, where the generated output $T_\theta(\mathbf{z})$ only models a subset of the training examples.

To address the issue of mode collapse, an alternative method Implicit Maximum Likelihood Estimation (IMLE)~\cite{li2018implicit} has been introduced. While IMLE, like GANs, uses a generator, it differs from GANs by using an alternative objective. The IMLE objective   ensures that \emph{each} training data point has similar generated samples, thereby encouraging coverage of all the modes of the training data.

The IMLE objective is given as follows, where $d(\cdot,\cdot)$ is a distance metric:
\begin{align}
    \theta_{\text{IMLE}}
    &= \argmin_\theta \mathbb{E}_{z_1,...,z_m \sim \mathcal{N}(0, I)}  \left [\sum_{i=1}^n \operatorname{min}\limits_{j\in [m]} d\left( \mathbf{x}_i, T_\theta(\mathbf{z}_j)\right) \right ]
\label{eqn:imle}
\end{align}

Here $m$ denotes the number of samples and $n$ denotes the number of data points. In simple terms, the IMLE objective first draws $m$ samples $\mathbf{z}_{j}$ from the standard Gaussian distribution and transforms them into the image space using the function $T_{\theta}$. From these pool of samples in the image space, for each data point $x_{i}$, IMLE selects a sample that is \textit{closest} to the data point in some distance metric $d(\cdot,\cdot)$. This operation can be done efficiently due to advances in high-dimensional nearest neighbour search \cite{li2017fast}. Note that the number of samples $m$ must at least be equal to the number of data points $n$, since otherwise by the pigeonhole principle, some samples would be picked by multiple data points. In practice, we find that setting $m$ to be a multiplicative factor (like 10 or 20) times larger than $n$ works the best.

\subsection{Observation} \label{subsec:motivation}

In the existing IMLE-based methods, we observe that the distributions of the latent codes used for training the objective differs from the distribution of latent encountered at test time. Consider an illustrative example where the latent space is two dimensional. We train a simple generative model using IMLE on two dimensional toy dataset. The latent codes used for training over the course of training are illustrated in Figure~\ref{fig-banner-latent-imle}. 
We notice that for the latent codes belonging to the same data point (denoted by the same colour) form well-separated tight bands in the latent space. We also observe that there are large gaps between these bands, indicating that these segments of the latent space are consistently overlooked during training. Since at test time we sample from the same standard normal distribution, these unsupervised segments in the latent space have arbitrary outputs, which result in bad samples. We term this phenomenon the ``\textit{misalignment issue}.''

\subsection{Analysis of the Misalignment Issue}

In this section, we will explore why the phenomenon mentioned above occurs. Let us clarify the notation used in subsequent sections: $\mathbf{x}_{i}$ denotes the $i^{\text{th}}$ data point, $d(\cdot,\cdot)$ denotes the distance function and $T_{\theta}(\mathbf{z}_{j})$ denote the $j^{\text{th}}$ sample where $\mathbf{z}_j \sim \mathcal{N}(0, I)$. 

Let us define a random variable, $D_{ij}$ to denote the distance of $i^{\text{th}}$ data point to the $j^{\text{th}}$ sample. We can also define another random variable $D_{i}^{*}$ to denote the distance of $i^{\text{th}}$ data point to the sample \emph{closest} to it. Further, let $F_{{D}_{ij}}$ and $F_{{D}^{*}_{i}}$ be the CDF of ${D}_{ij}$ and ${D}^{*}_{i}$ respectively. Let $f_{{D}_{ij}}$ and $f_{{D}^{*}_{i}}$ be the PDF of ${D}_{ij}$ and ${D}^{*}_{i}$ respectively. 
Now we will relate the CDF of the distances between a data point and its selected latent code, $F_{{D}^{*}_{i}}$ to the CDF of the distances between the same data point and a random latent code $F_{{D}_{ij}}$:

\begin{align}
    F_{{D}^{*}_{i}}(t) &= \Pr({D}^{*}_{i} \leq t) = 1 - \Pr({D}^{*}_{i} > t) \nonumber  \\
    &= 1 - \Pr({D}_{ij} > t, \forall j\in [m] ) & \left(\text{Def. of } {D}_{ij}\right) \nonumber \\
    &= 1 - \prod_{j=1}^{m} \Pr({D}_{ij} > t) \nonumber \\
    &= 1 - \left(\Pr({D}_{i1} > t)\right)^m & \left( {D}_{ij}\text{ are i.i.d}\right) \label{eqn:equal-in-dist} \\
    &= 1 - \left(1 - F_{{D}_{i1}}(t)\right)^m \label{eqn:fstar-vs-f}
\end{align}

Note that Equation~\ref{eqn:equal-in-dist} is true because each $\mathbf{{z}}_j$ is drawn independently from the same probability distribution which makes ${D}_{i1}, {D}_{i2} \cdots {D}_{im}$ identical in distribution for a particular data point $\mathbf{x}_i$. 

\begin{figure*}[!t]
  \captionsetup[subfigure]{labelformat=parens}
  \begin{subfigure}[t]{0.32\textwidth}
    \centering
    \includegraphics[width=\linewidth]{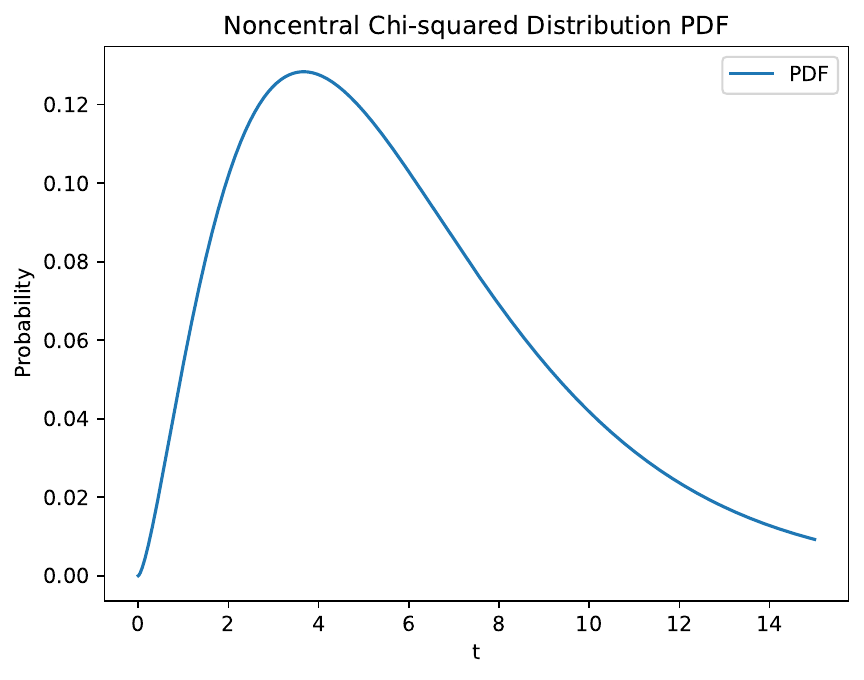}
    \caption{{PDF of example distribution}}
    \label{fig-theory-a}
  \end{subfigure}
  \hfill
  \begin{subfigure}[t]{0.32\textwidth}
    \centering
    \includegraphics[width=\linewidth]{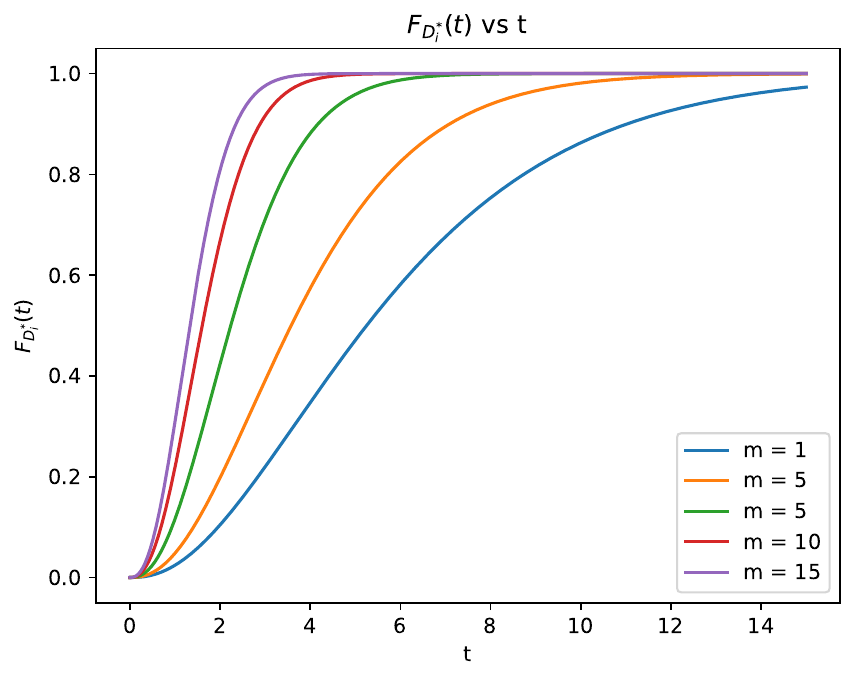}
    \caption{{$F_{{D}^{*}_{i}}(t)$ vs $t$ for different values of $m$}}
    \label{fig-theory-b}
  \end{subfigure}
  \hfill
  \begin{subfigure}[t]{0.32\textwidth}
    \centering
    \includegraphics[width=\linewidth]{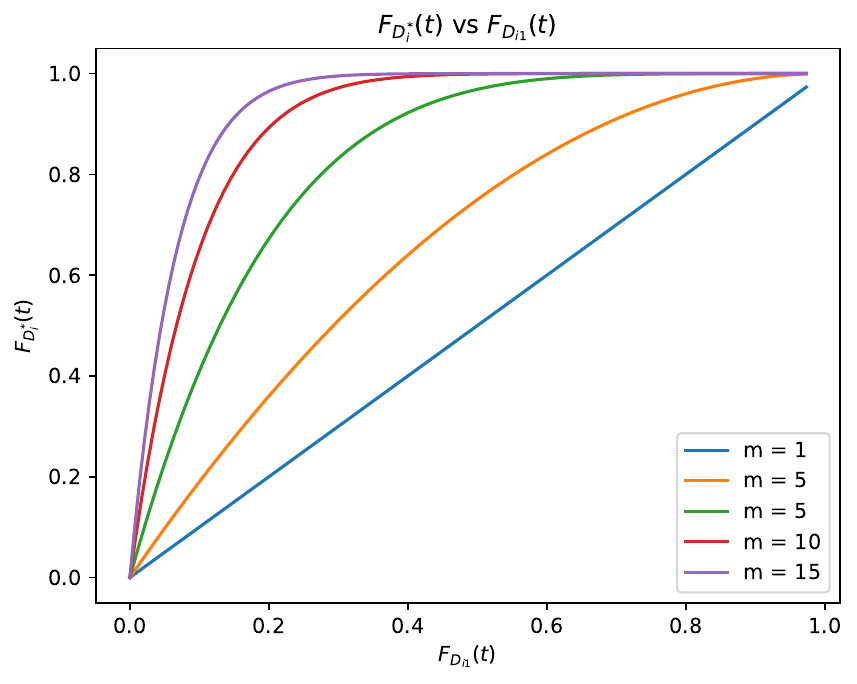}
    \caption{{$F_{{D}^{*}_{i}}(t)$ vs $F_{{D}_{i1}}(t)$ for different values of $m$}}
    \label{fig-theory-c}
  \end{subfigure}

  \caption{Illustrative figure for demonstrating the behaviour of $F_{{D}^{*}_{i}}(t)$ and $F_{{D}_{i1}}(t)$ using Noncentral Chi-squared distribution as the example distribution.}
  \label{fig-theory-non-central}
\end{figure*}

Now we can try to justify our observations from Figure~\ref{fig-banner-latent-imle}. Equation~\ref{eqn:fstar-vs-f} shows us how the CDF of the distance of the selected latent code (used in training) differs significantly from the CDF of distance of a random latent code (encountered at test time). This shows that the distance of the sample we choose for training is typically lower than the distance of the sample at testing.
We can obtain a deeper understanding by analyzing the plots in Figure~\ref{fig-theory-non-central} for an example distribution. Notice that $\forall m > 1, F_{{D}^{*}_{i}}(t) > F_{{D}_{i1}}(t) $.
We can observe the following from Equation~\ref{eqn:fstar-vs-f} and the aforementioned plots: 

\begin{enumerate}
    \item The $f_{{D}^{*}_{i}}$ is skewed towards the origin compared to $f_{{D}_{ij}}$. This is intuitive because the latent codes are selected by the $\operatorname{min}$ operation and so their distance to their respective data point would be less than that of a random sample.
    \item The skew towards the data point increases as $m$ increases. This observation will be important shortly.
\end{enumerate}

We can also compute the PDF $f_{{D}^{*}_{i}}$ in terms of $f_{{D}_{i1}}$ by differentiating the CDF $F_{{D}^{*}_{i}}(t)$ as follows:

\begin{align}
    f_{{D}^{*}_{i}}(t) &= \frac{\mathrm{d} F_{{D}^{*}_{i}}(t)}{\mathrm{d}t}
    = m \left(1 - F_{{D}_{i1}}(t)\right)^{m-1} f_{{D}_{i1}}(t) \label{eqn:p-pdf}
\end{align} 

\subsection{Solving the Misalignment Issue}

Now that we know the reason behind the misalignment between latent codes used at training and testing, we wish to come up with a method to mitigate this phenomenon. Recall that $D_{ij}$ and $D^{*}_{i}$ are determined by the distance function $d(\cdot,\cdot)$, neural network $T_{\theta}$ and the prior distribution. For a given generative modelling task, it is not trivial to change the $d(\cdot,\cdot)$ or $T_{\theta}$. Hence, in this paper, we aim to change the prior distribution used at training such that the distribution of latent codes at training time closely match with the distribution of latent codes (drawn from the standard normal distribution) encountered at test time. Notice that the IMLE objective (Equation~\ref{eqn:imle}) allows us to sample from the prior without knowing the closed form expression for its probability density function. This allows us the flexibility of choosing a non-analytical prior distribution. To distinguish from $\mathbf{z}_j \sim \mathcal{N}(0, I)$, we will use $\Tilde{\mathbf{z}}$ to denote the latent codes drawn from our desired target distribution $\mathcal{P}$.
Identical to the previous section, let us define random variable $\Tilde{D}_{ij} = d\left( \mathbf{x}_i, T_\theta(\mathbf{\Tilde{z}}_j)\right)$ to denote the distance between data point $\mathbf{x}_i$ from a random sample $T_\theta(\mathbf{\Tilde{z}}_j)$. Similarly, we define $\Tilde{D}^{*}_{i} = \operatorname{min}\limits_{j\in [m]} \Tilde{D}_{ij}$. Similar to the section above, let $F_{\Tilde{D}_{ij}}$ and $F_{\Tilde{D}^{*}_{i}}$ be the CDF of $\Tilde{D}_{ij}$ and $\Tilde{D}^{*}_{i}$ respectively. Then $f_{\Tilde{D}_{ij}}$ and $f_{\Tilde{D}^{*}_{i}}$ would be the PDF of $\Tilde{D}_{ij}$ and $\Tilde{D}^{*}_{i}$ respectively. 

\begin{algorithm}[t]
\footnotesize
\caption{\label{alg:eps-IMLE}RS-IMLE Procedure}
\begin{algorithmic}[1]
\Require The set of inputs $\left\{ \mathbf{x}_{i}\right\} _{i=1}^{n}$, radius $\epsilon$ 
\State Initialize the parameters $\theta$ of the generator $T_\theta$

\For{$k = 1$ \textbf{to} $K$}
    \State Draw latent codes $Z \gets \mathbf{z}_1,...,\mathbf{z}_m$ from $\mathcal{N}(0, \mathbf{I})$
    \State \begin{varwidth}[t]{\linewidth}
      Compute $\Tilde{Z} \gets \mathbf{\Tilde{z}}_1,...,\mathbf{\Tilde{z}}_p$ from $Z$ such that ~\par
        \hskip\algorithmicindent $d\left( \mathbf{x}_i, T_\theta(\mathbf{\Tilde{z}}_j)\right) \geq \epsilon, \quad \forall \mathbf{\Tilde{z}}_j \in \Tilde{Z}, i \in [n]$ \par
      \end{varwidth}
    \State $\sigma(i) \gets \arg \min_{j \in [m]} d(\mathbf{x}_{i}, T_{\theta}(\mathbf{\Tilde{z}_j})), \quad \forall i \in [n]$
    \For{$l = 1$ \textbf{to} $L$}
        \State Pick a random batch $S \subseteq [n]$
        \State $\theta \gets \theta - \eta \nabla_{\theta}\left(\sum_{i \in S}d\left(\mathbf{x}_{i}, T_{\theta}\left(\mathbf{\Tilde{z}}_{\sigma(i)}\right)\right)\right) / |S|$
    \EndFor

\EndFor
\State \Return $\theta$
\end{algorithmic}
\end{algorithm}

\subsubsection{Designing the target prior}

Now, we discuss the desired properties for our target prior. Recall that the misalignment issue is mitigated as the number of samples, denoted by $m$, decreases. In order to differentiate between the number of samples of different priors, we use the notation $m'$ to denote the number of samples for the objective using the Gaussian prior. As hinted in the previous section, one way to avoid the misalignment issue is to set $m'$ to low values. However we cannot directly use a Gaussian prior with arbitrary low values of $m'$ as our target prior. This is because having too few samples to choose from would cause many data points to pick the same sample as their nearest neighbour. 
Since the objective function tries to pull the nearest sample toward each data point, pulling the same sample towards different data points creates conflicting supervision signals, leading to slow convergence or even no learning, especially when target data points for the same sample lie in opposite directions.
Hence when using a Gaussian prior, we need to pick $m'$ large enough to allow convergence and yet small enough such that the misalignment issue does not affect test time performance. In our case, we are trying to design a new distribution that solves the misalignment issue by having desirable properties of an ideal distribution. The ideal distribution is a Gaussian prior with $m'$ set to the lowest possible value, which is $m' = n$.

To this end, we can choose a prior distribution $\mathcal{P}$ that matches the ideal prior distribution. Similar to the analysis till Equation~\ref{eqn:p-pdf}, we derive the PDF of $\mathcal{P}$. We get $f_{\Tilde{D}^{*}_{i}}(t) = m \left(1 - F_{\Tilde{D}_{i1}}(t)\right)^{m-1} f_{\Tilde{D}_{i1}}(t)$. Equating this PDF of $\mathcal{P}$ to the PDF of the ideal distribution gives:

\begin{align}
    & m \left(1 - F_{\Tilde{D}_{i1}}(t)\right)^{m-1} f_{\Tilde{D}_{i1}}(t) = n \left(1 - F_{{D}_{i1}}(t)\right)^{n-1} f_{{D}_{i1}}(t) \nonumber \\
    &\implies f_{\Tilde{D}_{i1}}(t) = \frac{n}{m} \frac{\left(1 - F_{{D}_{i1}}(t)\right)^{n-1}}{\left(1 - F_{\Tilde{D}_{i1}}(t)\right)^{m-1}} f_{{D}_{i1}}(t)  \label{eqn:pre-unnorm-f-tilde}
\end{align}
We introduce $\phi(t) = \frac{n}{m} \frac{\left(1 - F_{{D}_{i1}}(t)\right)^{n-1}}{\left(1 - F_{\Tilde{D}_{i1}}(t)\right)^{m-1}}$  to simplify notation. Hence, we can write Equation \ref{eqn:pre-unnorm-f-tilde} as:

\begin{align}
    f_{\Tilde{D}_{i1}}(t) = \phi(t) f_{{D}_{i1}}(t) \label{eqn:unnorm-f-tilde}
\end{align}

\subsubsection{Rejection sampling}

We have expressed our target prior $\mathcal{P}$ in terms of distribution from which we can easily sample. Now, we can use rejection sampling to sample from our target prior $\mathcal{P}$. \\
To be concrete: $f_{\Tilde{D}_{i1}}(t)$ is our target distribution and $f_{{D}_{i1}}(t)$ acts as our proposal distribution, since we can sample from the standard Gaussian easily. In order to ensure that the acceptance ratio is bounded, we introduce a constant $c$  associated with truncating $F_{{D}_{i1}}(t)$. We discuss these technical details in the appendix. \\
We can write the acceptance ratio in the standard rejection sampling notation: $\frac{f_{\Tilde{D}_{i1}}(t)}{M f_{{D}_{i1}}(t)} = \frac{c \phi(t)}{M}$ \\
Here, $M$ is the scaling factor associated with rejection sampling. We approximate the function above using a step function. The step needs to happen at $t$ where $F_{\tilde{D}_{i1}}(t)$ gets close to 1. Instead of trying to estimate $F_{\tilde{D}_{i1}}(t)$, we instead use a hyperparameter $\epsilon$ to represent where $F_{\tilde{D}_{i1}}(t)$ gets close to 1. We find the value of this hyperparameter $\epsilon$ by cross-validation. 

The final procedure simplifies to this: we sample ${\mathbf{z}} \sim \mathcal{N}(0, I)$. If $t = d\left( \mathbf{x}_i, T_\theta(\mathbf{{z}})\right) < \epsilon$, we reject the sample; otherwise, we accept it. Since the sampling procedure is based on rejection sampling, we call our method \emph{RS-IMLE}. The resulting RS-IMLE procedure is included in Algorithm-\ref{alg:eps-IMLE}.

\begin{figure*}[!t]
  \captionsetup[subfigure]{labelformat=parens}
  \begin{subfigure}[t]{0.45\textwidth}
    \centering
    \includegraphics[width=\linewidth]{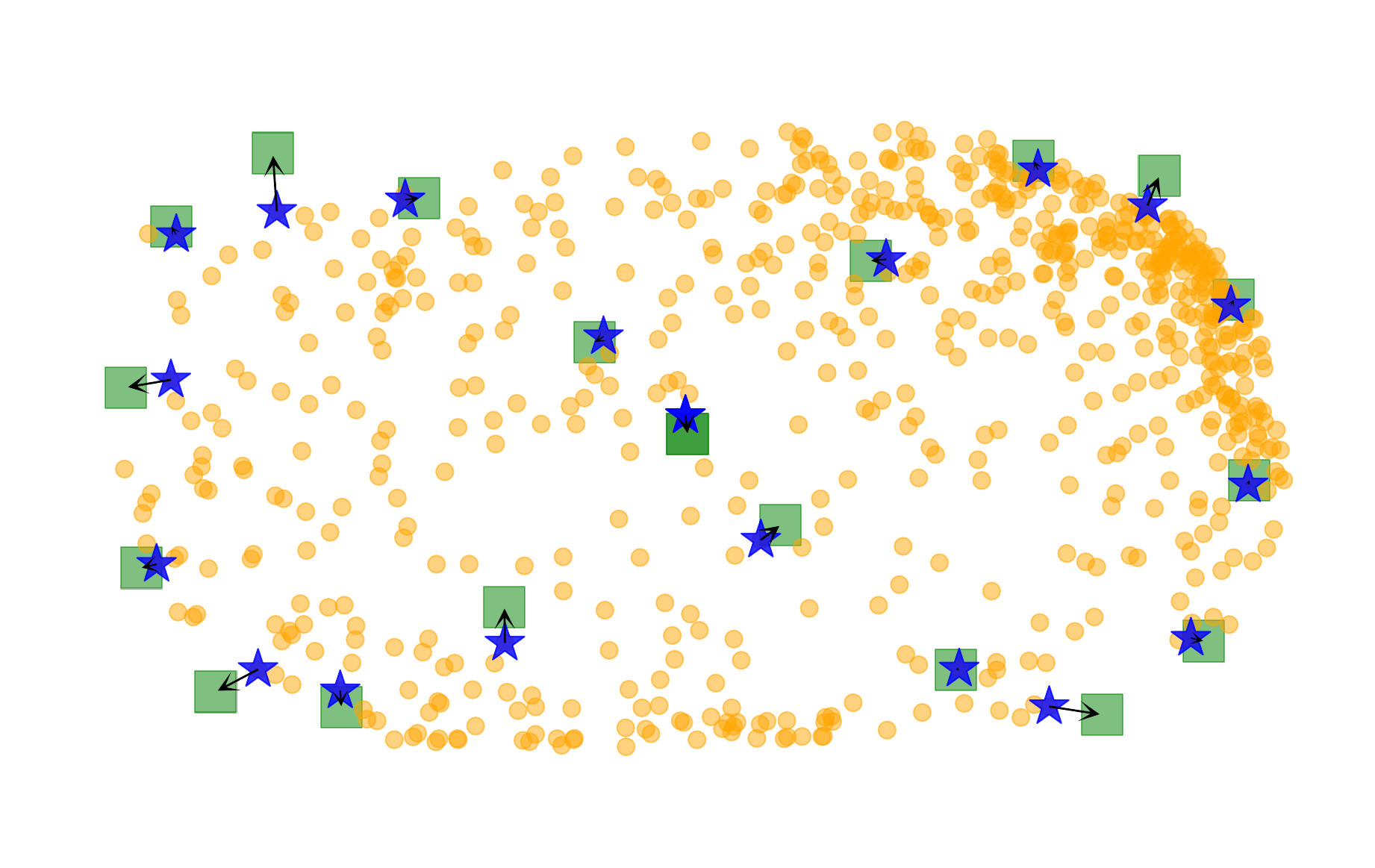}
    \caption{{IMLE after epoch 100}}
    \label{fig-comparision-imle:100}
  \end{subfigure}
  \hfill
  \begin{subfigure}[t]{0.45\textwidth}
    \centering
    \includegraphics[width=\linewidth]{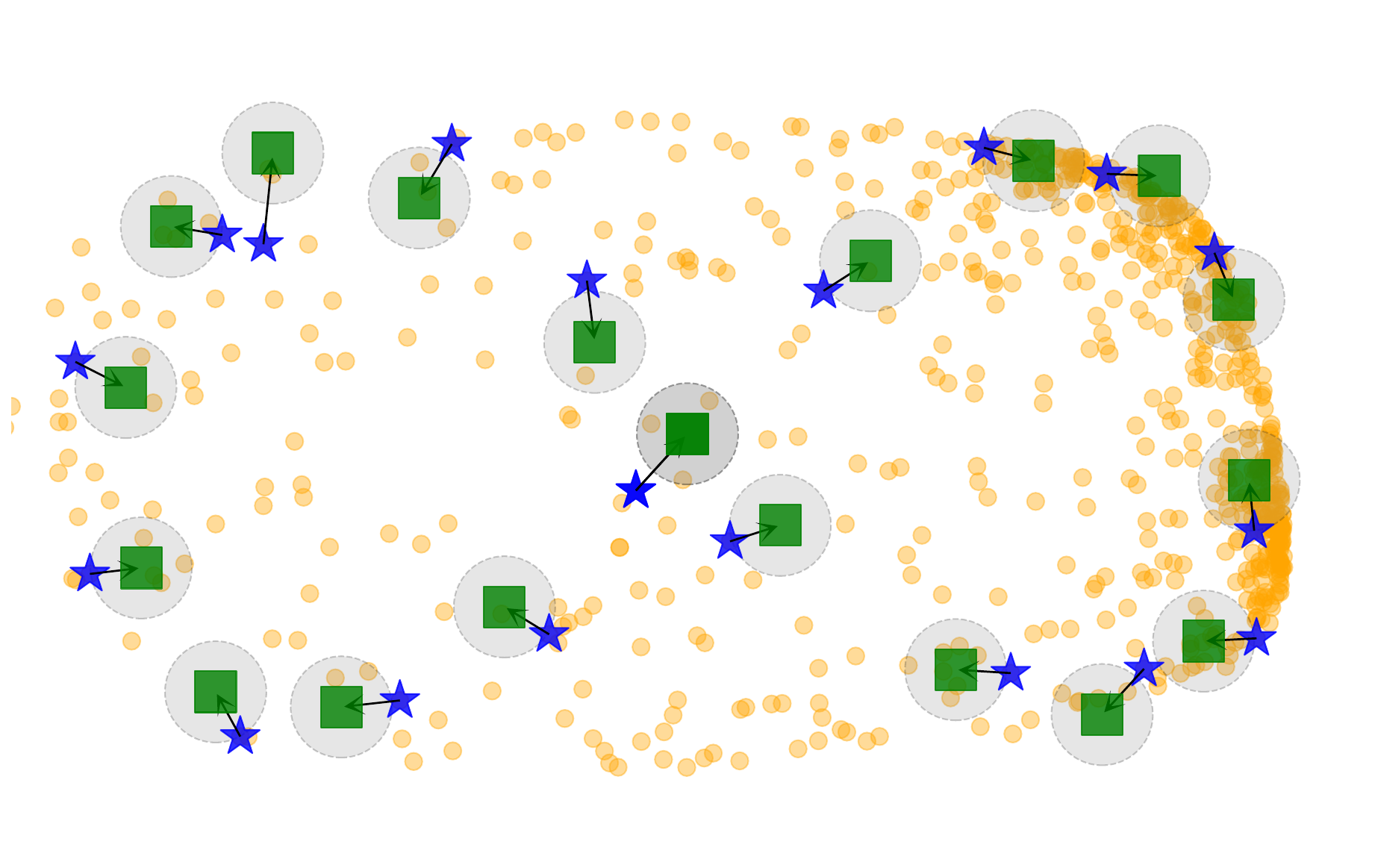}
    \caption{{RS-IMLE after epoch 100}}
    \label{fig-comparision-eps-imle:100}
  \end{subfigure}

  \begin{subfigure}[t]{0.45\textwidth}
    \centering
    \includegraphics[width=\linewidth]{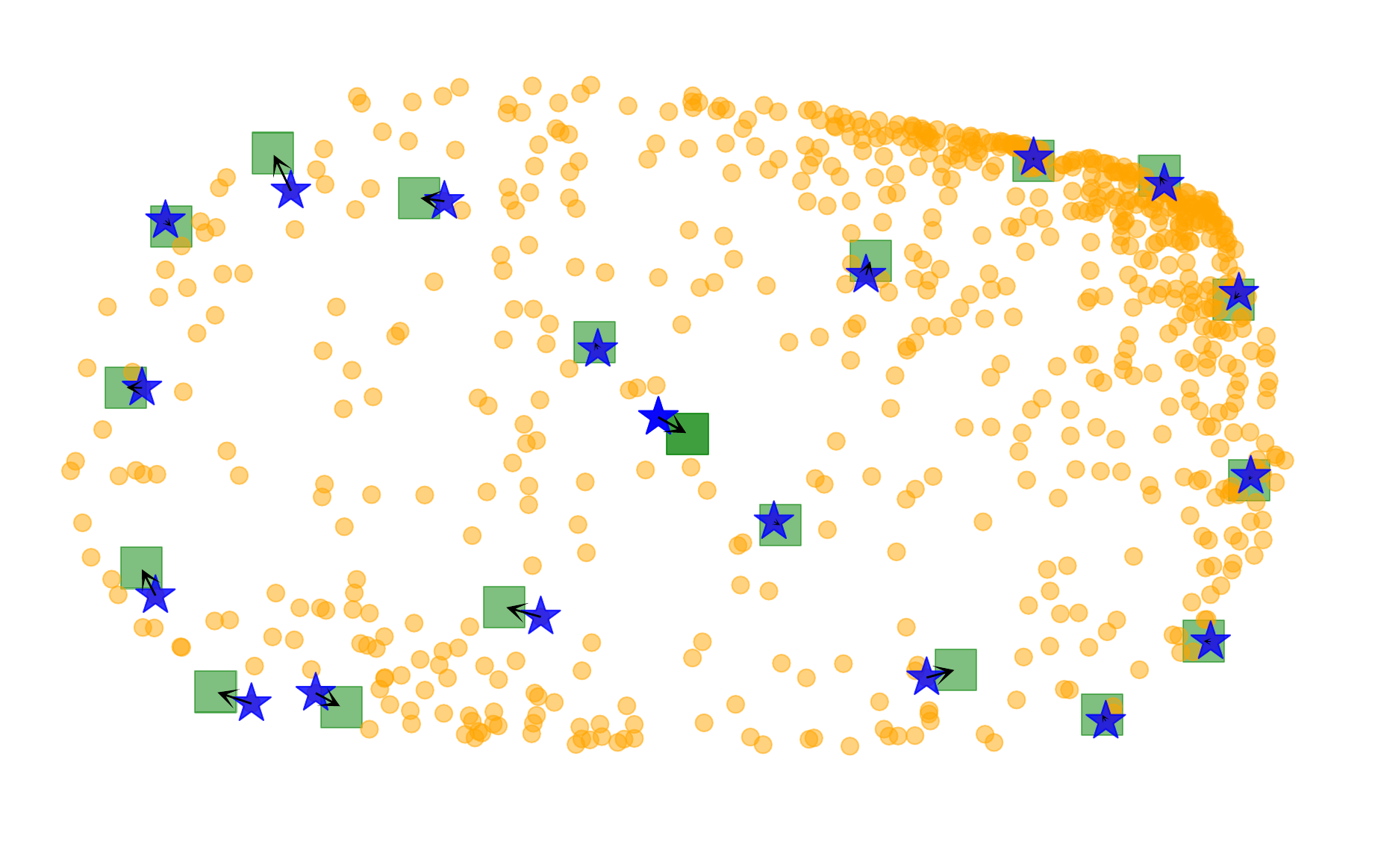}
    \caption{{IMLE after epoch 500} }
    \label{fig-comparision-imle:500}
  \end{subfigure}
  \hfill
  \begin{subfigure}[t]{0.45\textwidth}
    \centering
    \includegraphics[width=\linewidth]{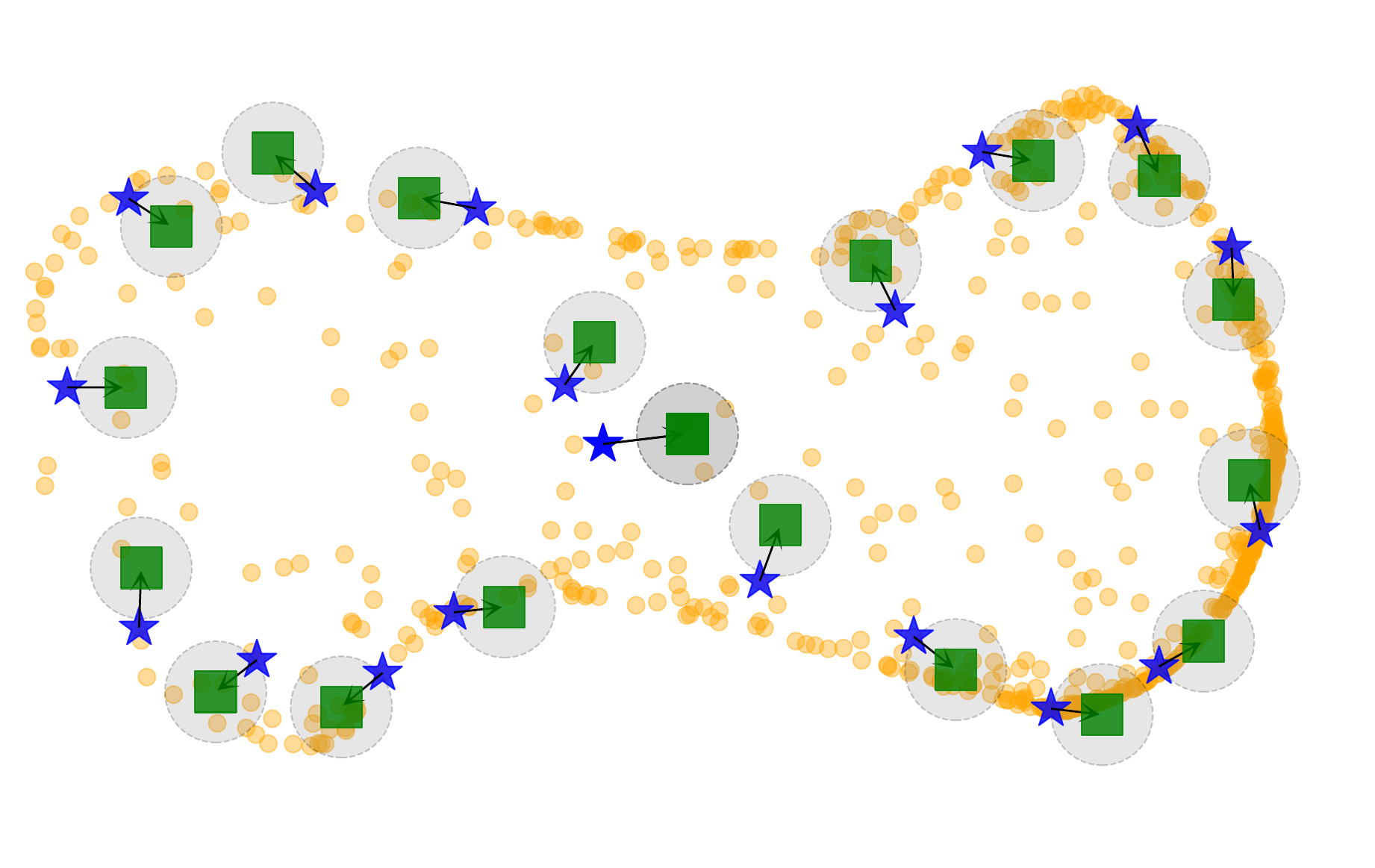}
    \caption{{RS-IMLE after epoch 500 }}
    \label{fig-comparision-eps-imle:500}
  \end{subfigure}

  \begin{subfigure}[t]{0.45\textwidth}
    \centering
    \includegraphics[width=\linewidth]{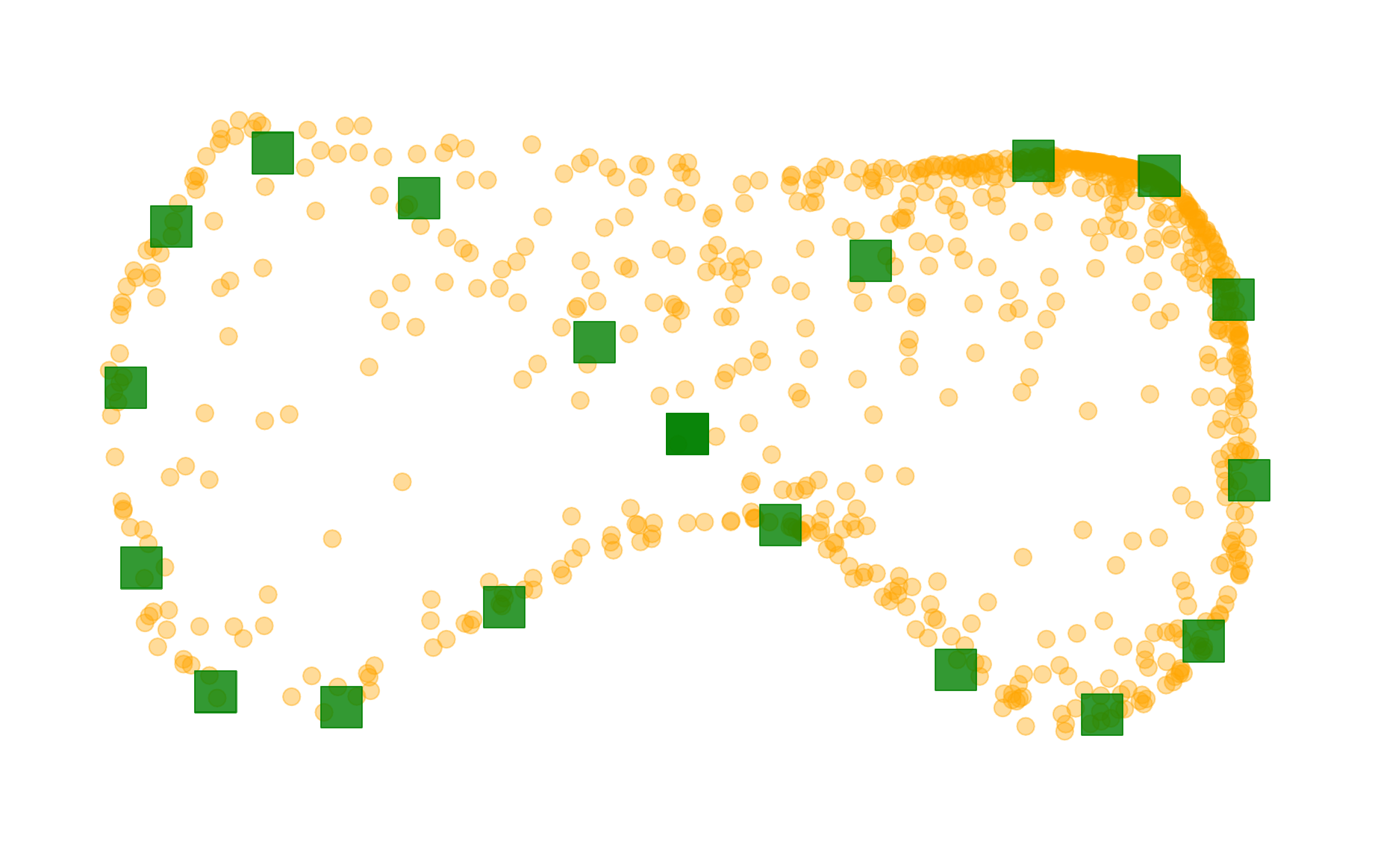}
    \caption{{IMLE after 2000 epochs}}
    \label{fig-comparision-imle:final}
  \end{subfigure}
  \hfill
  \begin{subfigure}[t]{0.45\textwidth}
    \centering
    \includegraphics[width=\linewidth]{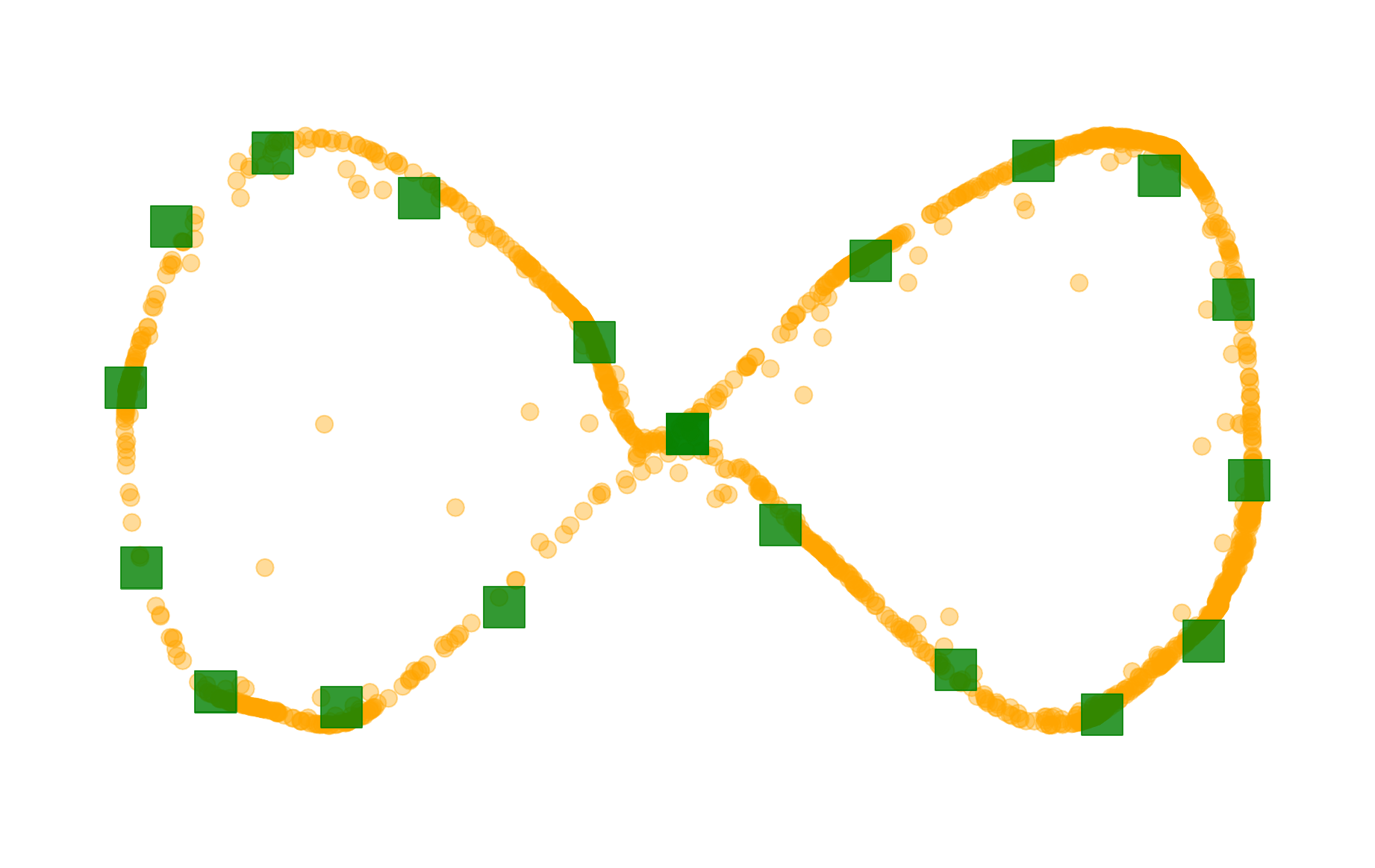}
    \caption{{RS-IMLE after 2000 epochs}}
    \label{fig-comparision-eps-imle:final}
  \end{subfigure}

  \caption{Comparison between IMLE and RS-IMLE for 2D toy problem. Data points are denoted by ${\color{ForestGreen}\mdblksquare}$ and samples are denoted by ${\color{circlecolor}\mdblkcircle}$. Samples picked as nearest neighbours are denoted by $\color{blue} \bigstar $.}
  \label{fig-comparision}
\end{figure*}

\subsection{Intuitive Interpretation of the Algorithm Behaviour}

Prior to proceeding to the implementation details, gaining a gradient-based understanding of our new objective would provide valuable insight. 
Let us revisit the vanilla IMLE objective (Equation~\ref{eqn:imle}) again. As training progresses, we expect the loss to reduce such that:
$\forall x_i, \quad \mathbb{E}_{z_1,...,z_m \sim \mathcal{N}(0, I)} \operatorname{min} \limits_{j\in [m]} d\left( \mathbf{x}_i, T_\theta(\mathbf{z}_j)\right) \rightarrow 0$. 
As the objective approaches convergence, the loss will decrease, resulting in gradients with lower magnitude. This causes smaller updates to the model parameters during training, leading to slower progress. Note that this loss is with respect to the \emph{closest} sample to each data point. Thus it can be the case that even after a lot of training, although the \emph{closest} sample are pretty close to their respective data point, the rest of samples are pretty far away. 

\begin{align}
    \theta_{\text{RS-IMLE}}
    &= \argmin_\theta \mathbb{E}_{z_1,...,z_m \sim \mathcal{P}}  \left [\sum_{i=1}^n \operatorname{min}\limits_{j\in [m]} d\left( \mathbf{x}_i, T_\theta(\mathbf{z}_j)\right) \right ]
\label{eqn:eps-imle}
\end{align}

Consider our objective in Equation~\ref{eqn:eps-imle} and recall that we have constructed the probability distribution $\mathcal{P}$ such that $\forall \mathbf{x}_i \quad d\left( \mathbf{x}_i, T_\theta(\mathbf{{\mathbf{\Tilde{z}}}})\right) \geq \epsilon$, where $\mathbf{\Tilde{z}} \sim \mathcal{P}$. In other words, all samples we obtain by using the prior $\mathcal{P}$ are guaranteed to be $\epsilon$-distance away from all data points. This ensures that the loss per data point is always greater that $\epsilon$. The approach can be interpreted as ignoring the samples that are already close to some data point and instead training on challenging, non-trivial samples.

To compare the sampling behavior of the vanilla IMLE and the proposed RS-IMLE, we trained two models on a 2D toy problem as illustrated in Figure~\ref{fig-comparision}. The first model uses the vanilla IMLE objective (Equation~\ref{eqn:imle}), while the second model is trained with our proposed RS-IMLE objective (Equation~\ref{eqn:eps-imle}).

At the initial stage of training, both the methods learn similar distributions, indicated by the straight line of orange dots (${\color{circlecolor}\mdblkcircle}$) in Figure-\ref{fig-comparision-imle:100},\ref{fig-comparision-eps-imle:100}. Our method first removes all the samples that fall within an $\epsilon$ distance from any data point before doing the nearest neighbour search. We illustrate this in Figure-\ref{fig-comparision-eps-imle:100}, where samples that lie within the gray circles are not considered for the nearest neighbour search.

As training progresses, we notice that for both the algorithms' samples move closer to the ground truth data points. However, for vanilla IMLE, we observe that for many data points the sample picked after the nearest neighbour search is already close to the ground truth. In this case, the loss associated with these data points would be low (indicated by the short length of arrow in the Figure-\ref{fig-comparision-imle:500}). As a result, the model trained by the vanilla IMLE objective does not learn anything significantly novel. In our proposed method (Figure-\ref{fig-comparision-eps-imle:500}), each data point selects a sample that is at least $\epsilon$ distance away from it. This  ensures that the loss for each data point is always sufficiently high (indicated by the long arrows), resulting in meaningful updates to the model parameters.

\subsection{Implementation Details}

Note that computing the distance of each sample with each data point is computationally expensive. Suppose we have $n$ data points in $\mathbb{R}^d$ and $m$ samples, calculating the distance between each pair has a time complexity of $\mathcal{O}(mnd)$. To get around this issue, we leverage a fast k-nearest neighbor search method, DCI~\cite{li2017fast}. This method reduces the runtime of a single query from linear to sublinear in $m$, enabling us to efficiently \emph{filter out} all the samples that are within an $\epsilon$ distance from any data point. Subsequently, from this filtered pool of remaining samples, we select, for each data point, the sample that is closest to it.

In order to reduce the search time complexity further, we can project the training data to a lower dimensional subspace (which would decrease $d$). We use this while implementing the procedure for image synthesis task: we first flatten each image and project it to a lower dimension by using random projection. We normalize these projected vectors before using them for nearest neighbour search. 

\section{Experiments}

\begin{figure*}[t]
  \captionsetup[subfigure]{labelformat=parens}  
  \begin{subfigure}[t]{0.16\textwidth}
    \centering
    FastGAN
  \end{subfigure}
  \begin{subfigure}[t]{0.16\textwidth}
    \centering
    FakeCLR
  \end{subfigure}
  \begin{subfigure}[t]{0.16\textwidth}
    \centering
    FreGAN
  \end{subfigure}
  \begin{subfigure}[t]{0.16\textwidth}
    \centering
    ReGAN
  \end{subfigure}
  \begin{subfigure}[t]{0.16\textwidth}
    \centering
    AdaIMLE
  \end{subfigure}
  \begin{subfigure}[t]{0.16\textwidth}
    \centering
    RS-IMLE
  \end{subfigure}

  \begin{subfigure}[t]{0.16\textwidth}
    \centering
    \includegraphics[width=\linewidth]{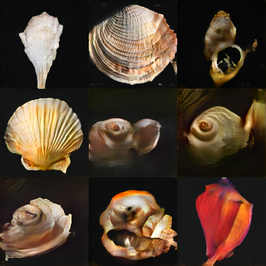}
  \end{subfigure}
  \begin{subfigure}[t]{0.16\textwidth}
    \centering
    \includegraphics[width=\linewidth]{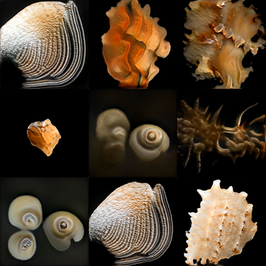}
  \end{subfigure}
  \begin{subfigure}[t]{0.16\textwidth}
    \centering
    \includegraphics[width=\linewidth]{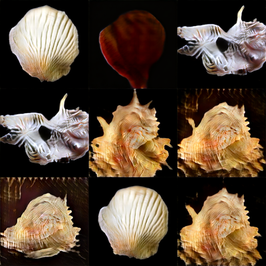}
  \end{subfigure}
  \begin{subfigure}[t]{0.16\textwidth}
    \centering
    \includegraphics[width=\linewidth]{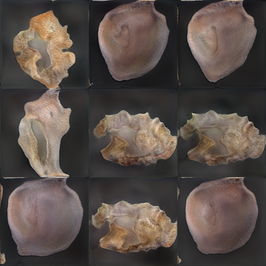}
  \end{subfigure}
  \begin{subfigure}[t]{0.16\textwidth}
    \centering
    \includegraphics[width=\linewidth]{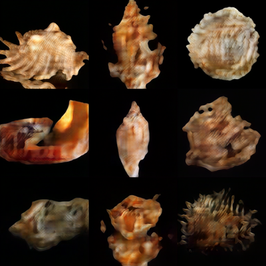}
  \end{subfigure}
  \begin{subfigure}[t]{0.16\textwidth}
    \centering
    \includegraphics[width=\linewidth]{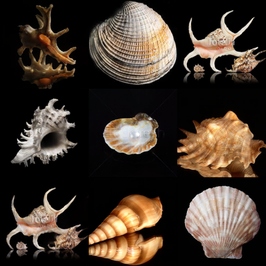}
  \end{subfigure}
  \begin{subfigure}[t]{0.16\textwidth}
    \centering
    \includegraphics[width=\linewidth]{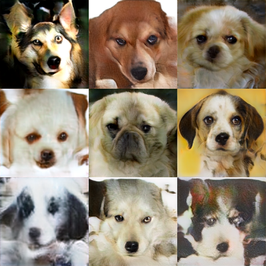}
  \end{subfigure}
  \begin{subfigure}[t]{0.16\textwidth}
    \centering
    \includegraphics[width=\linewidth]{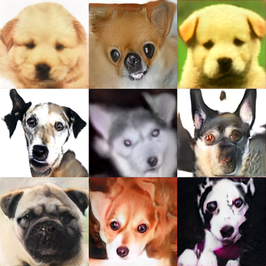}
  \end{subfigure}
  \begin{subfigure}[t]{0.16\textwidth}
    \centering
    \includegraphics[width=\linewidth]{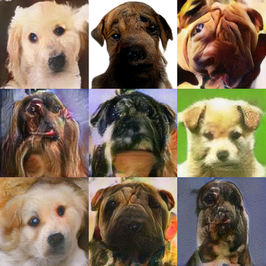}
  \end{subfigure}
  \begin{subfigure}[t]{0.16\textwidth}
    \centering
    \includegraphics[width=\linewidth]{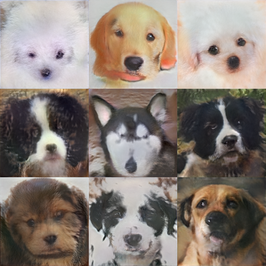}
  \end{subfigure}
  \begin{subfigure}[t]{0.16\textwidth}
    \centering
    \includegraphics[width=\linewidth]{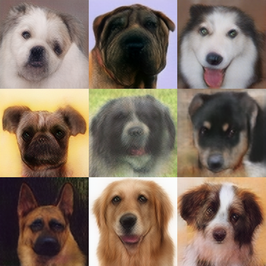}
  \end{subfigure}
  \begin{subfigure}[t]{0.16\textwidth}
    \centering
    \includegraphics[width=\linewidth]{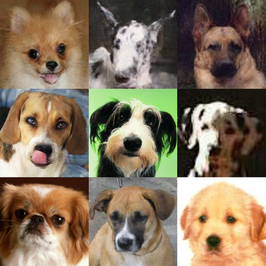}
  \end{subfigure}
  \caption{\textbf{Qualitative comparison between our method and baselines.} While analyzing the images, look for the sharpness of each image and diversity in the content of all images for a method.}
  \label{fig-random-samples}
\end{figure*}

\noindent \textbf{Datasets} We assess our method and the baseline approaches across a variety of datasets with $256 \times 256$ resolution. These datasets include Animal-Face Dog~\cite{dataset-cg}, Animal-Face Cat~\cite{dataset-cg}, Obama\cite{Zhao2020DifferentiableAF}, Panda\cite{Zhao2020DifferentiableAF}, Grumpy-cat~\cite{Zhao2020DifferentiableAF}, Anime \cite{DBLP:fastgan}, Shells \cite{DBLP:fastgan}, Skulls \cite{DBLP:fastgan} and a subset of Flickr-FaceHQ (FFHQ)~\cite{karras2019stylebased} which are standard datasets used in the few-shot learning literature.

\begin{table}[ht]
\centering
\setlength{\belowcaptionskip}{-1em}
\footnotesize
\begin{tabularx}{\textwidth}{xnnnnnns}
\toprule
\textbf{Dataset} & \textit{FastGAN} \cite{DBLP:fastgan} & \textit{FakeCLR} \cite{fakeclr} & \textit{FreGAN} \cite{yang2022FreGAN} & \textit{ReGAN} \cite{ReGAN} & \textit{AdaIMLE} \cite{adaptiveIMLE} & \textit{RS-IMLE (Ours)} & Imp. \% \\ \midrule
Obama      & 41.1  & 29.9  & 33.4  & 45.7  & 25.0  & \textbf{14.0} & 44.0 \\ 
Grumpy Cat & 26.6  & 20.6  & 24.9  & 27.3  & 19.1  & \textbf{11.5} & 39.8 \\ 
Panda      & 10.0  & 8.8   & 9.0   & 12.6  & 7.6   & \textbf{3.5}  & 54.0 \\ 
FFHQ-100   & 54.2  & 62.1  & 50.5  & 87.4  & 33.2  & \textbf{12.9} & 61.1 \\ 
Cat        & 35.1  & 27.4  & 31.0  & 42.1  & 24.9  & \textbf{15.9} & 36.1 \\ 
Dog        & 50.7  & 44.4  & 47.9  & 57.2  & 43.0  & \textbf{23.1} & 46.3 \\ 
Anime      & 69.8  & 77.7  & 59.8  & 110.8 & 65.8  & \textbf{35.8} & 45.6 \\ 
Skulls     & 109.6 & 106.5 & 163.3 & 130.7 & 81.9  & \textbf{51.1} & 37.6 \\ 
Shells     & 120.9 & 148.4 & 169.3 & 236.1 & 108.5 & \textbf{55.4} & 48.9 \\ 
\bottomrule
\end{tabularx}
\caption{We compute FID \cite{fid} between the real data and $5000$ randomly generated samples for all the methods. Lower is better.}
\label{tab:fid}
\end{table}

\noindent \textbf{Baselines} We compare our method to recent state-of-the-art few-shot image generation methods. These include FastGAN \cite{DBLP:fastgan}, FakeCLR \cite{fakeclr}, FreGAN \cite{yang2022FreGAN}, Re-GAN \cite{ReGAN} and AdaIMLE \cite{adaptiveIMLE}. 

\noindent \textbf{Evaluation Metrics} We employ Fréchet Inception Distance (FID)~\cite{fid} to assess the perceptual quality of the generated images. This involves randomly generating $5000$ images and calculating the FID between these generated samples and the real images for each dataset. Additionally, we evaluate the modelling accuracy and coverage by computing precision and recall for 1000 images using the metric defined by Kynkäänniemi et al. \cite{kynkaanniemi2019improved}. In image synthesis, precision refers to the model's capacity to generate images closely resembling the desired target or distribution. Recall measures the model's ability to encompass a wide array of diverse images within the target distribution. To ensures that the computed metrics have low variance, we generate many more samples than training images. 

\noindent \textbf{Network Architecture} We construct our generator network using decoder modules from VDVAE~\cite{vdvae}. More details about the network architecture can be found in the Appendix.

\subsection{Quantitative Results}

In Table~\ref{tab:fid}, we present the FID scores computed for all the datasets across different methods. Lower FID scores indicates that the distribution of generated images is closer to the distribution of real images. Our method performs significantly better compared to baselines. 

In Table~\ref{tab:pr}, we show the precision and recall scores. Our method has a near perfect precision (close to 1), while having a significantly higher recall compared to the baselines. In the few cases where our method is not the best, it is very close to the best metric.

\begin{table}[]
\setlength{\belowcaptionskip}{-1em}
\centering
\footnotesize
\begin{tabularx}{\linewidth}{xsnnnnnn}
\toprule
\textbf{Dataset}            &       & \textit{FastGAN} \cite{DBLP:fastgan} & \textit{FakeCLR} \cite{fakeclr} & \textit{FreGAN} \cite{yang2022FreGAN} & \textit{ReGAN} \cite{ReGAN} & \textit{AdaIMLE} \cite{adaptiveIMLE} & \textit{RS-IMLE (Ours)} \\ 
\midrule
\multirow{2}{*}{Obama}      & Prec. & 0.92             & 0.96    & 0.82    & 0.62   & \textbf{0.99}    & 0.98     \\ 
                            & Rec.  & 0.09             & 0.30    & 0.33    & 0.01   & 0.68    & \textbf{0.82}     \\ \hline
\multirow{2}{*}{Grumpy Cat} & Prec. & 0.91             & 0.97    & 0.90    & 0.78   & \textbf{0.97}    & 0.93     \\ 
                            & Rec.  & 0.13             & 0.39    & 0.23    & 0.04   & 0.72    & \textbf{0.95}     \\ \hline
\multirow{2}{*}{Panda}      & Prec. & 0.96             & 0.97    & 0.92    & 0.93   & 0.98    & \textbf{0.99}     \\ 
                            & Rec.  & 0.16             & 0.41    & 0.17    & 0.01   & 0.63    & \textbf{0.97}     \\ \hline
\multirow{2}{*}{FFHQ-100}   & Prec. & 0.91             & 0.71    & 0.86    & 0.39   & 0.99    & \textbf{1.00}     \\  
                            & Rec.  & 0.13             & 0.25    & 0.21    & 0.01   & 0.77    & \textbf{0.99}     \\ \hline
\multirow{2}{*}{Cat}        & Prec. & 0.97             & 0.99    & 0.95    & 0.90   & \textbf{0.98 }   & 0.96     \\  
                            & Rec.  & 0.08             & 0.55    & 0.31    & 0.15   & 0.86    & \textbf{0.98}     \\ \hline
\multirow{2}{*}{Dog}        & Prec. & 0.96             & 0.95    & 0.92    & 0.84   & 0.97    & \textbf{0.98 }    \\ 
                            & Rec.  & 0.19             & 0.34    & 0.20    & 0.10   & 0.61    & \textbf{0.94}     \\ \hline
\multirow{2}{*}{Anime}      & Prec. & 0.86             & 0.88    & 0.88    & 0.29   & 0.92    & \textbf{0.95}     \\ 
                            & Rec.  & 0.08             & 0.09    & 0.09    & 0.01   & 0.59    & \textbf{0.91}     \\ \hline
\multirow{2}{*}{Skulls}     & Prec. & 0.78             & 0.66    & 0.66    & 0.66   & 0.95    & \textbf{0.99 }    \\ 
                            & Rec.  & 0.03             & 0.01    & 0.01    & 0.01   & 0.32    & \textbf{0.65 }    \\ \hline
\multirow{2}{*}{Shells}     & Prec. & 0.92             & 0.87    & 0.87    & 0.50   & 0.97    & \textbf{0.98}     \\  
                            & Rec.  & 0.03             & 0.01    & 0.01    & 0.01   & \textbf{0.62}    & 0.59     \\ 
\bottomrule
\end{tabularx}%
\caption{We compute precision and recall \cite{precrecall} between the real data and $1000$ randomly generated samples for all the methods. Higher values are better.}
\label{tab:pr}
\end{table}

\begin{figure*}
  \captionsetup[subfigure]{labelformat=parens}
  
  \begin{subfigure}[t]{0.19\textwidth}
    \centering
    Query
  \end{subfigure}
  \begin{subfigure}[t]{0.19\textwidth}
    \centering
    RS-IMLE
  \end{subfigure}
  \begin{subfigure}[t]{0.19\textwidth}
    \centering
    Ada-IMLE
  \end{subfigure}
  \begin{subfigure}[t]{0.19\textwidth}
    \centering
    FastGAN
  \end{subfigure}
  \begin{subfigure}[t]{0.19\textwidth}
    \centering
    FreGAN
  \end{subfigure}

  \begin{subfigure}[t]{0.19\textwidth}
    \centering
    \includegraphics[width=\linewidth]{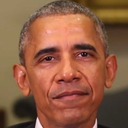}
    \caption{Obama}
    \label{fig-visual-recall:obama}
  \end{subfigure}
  \begin{subfigure}[t]{0.19\textwidth}
    \centering
    \includegraphics[width=\linewidth]{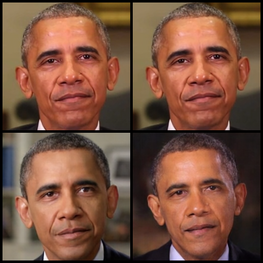}
  \end{subfigure}
  \begin{subfigure}[t]{0.19\textwidth}
    \centering
    \includegraphics[width=\linewidth]{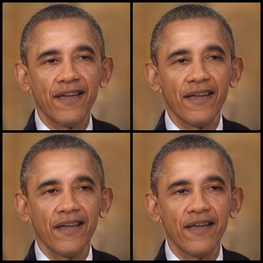}
  \end{subfigure}
  \begin{subfigure}[t]{0.19\textwidth}
    \centering
    \includegraphics[width=\linewidth]{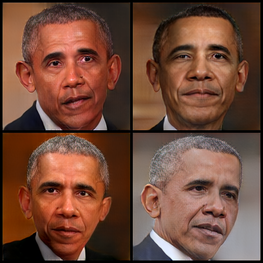}
  \end{subfigure}
  \begin{subfigure}[t]{0.19\textwidth}
    \centering
    \includegraphics[width=\linewidth]{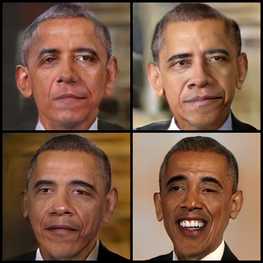}
  \end{subfigure}

  \begin{subfigure}[t]{0.19\textwidth}
    \centering
    \includegraphics[width=\linewidth]{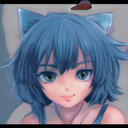}
    \caption{Anime}
    \label{fig-visual-recall:anime}
  \end{subfigure}
  \begin{subfigure}[t]{0.19\textwidth}
    \centering
    \includegraphics[width=\linewidth]{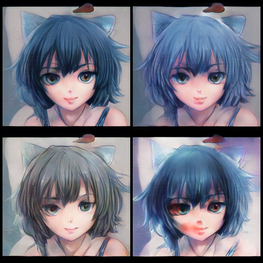}
  \end{subfigure}
  \begin{subfigure}[t]{0.19\textwidth}
    \centering
    \includegraphics[width=\linewidth]{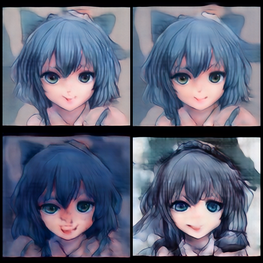}
  \end{subfigure}
  \begin{subfigure}[t]{0.19\textwidth}
    \centering
    \includegraphics[width=\linewidth]{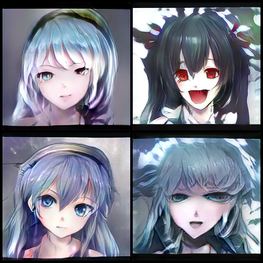}
  \end{subfigure}
  \begin{subfigure}[t]{0.19\textwidth}
    \centering
    \includegraphics[width=\linewidth]{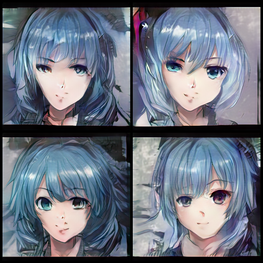}
  \end{subfigure}
  
  \begin{subfigure}[t]{0.19\textwidth}
    \centering
    \includegraphics[width=\linewidth]{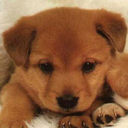}
    \caption{Dog}
    \label{fig-visual-recall:dog}
  \end{subfigure}
  \begin{subfigure}[t]{0.19\textwidth}
    \centering
    \includegraphics[width=\linewidth]{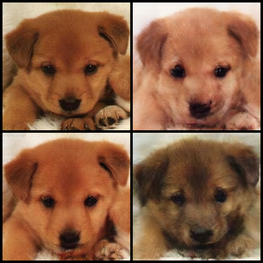}
  \end{subfigure}
  \begin{subfigure}[t]{0.19\textwidth}
    \centering
    \includegraphics[width=\linewidth]{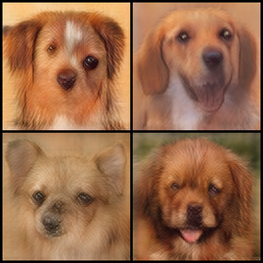}
  \end{subfigure}
  \begin{subfigure}[t]{0.19\textwidth}
    \centering
    \includegraphics[width=\linewidth]{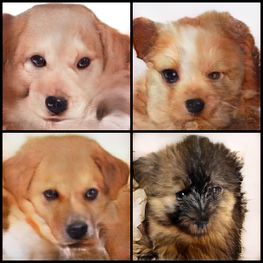}
  \end{subfigure}
  \begin{subfigure}[t]{0.19\textwidth}
    \centering
    \includegraphics[width=\linewidth]{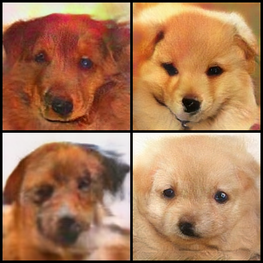}
  \end{subfigure}

  \begin{subfigure}[t]{0.19\textwidth}
    \centering
    \includegraphics[width=\linewidth]{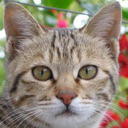}
    \caption{Cat}
    \label{fig-visual-recall:cat}
  \end{subfigure}
  \begin{subfigure}[t]{0.19\textwidth}
    \centering
    \includegraphics[width=\linewidth]{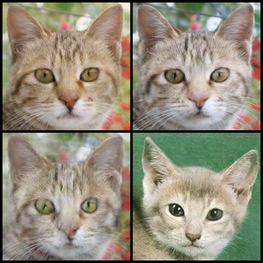}
  \end{subfigure}
  \begin{subfigure}[t]{0.19\textwidth}
    \centering
    \includegraphics[width=\linewidth]{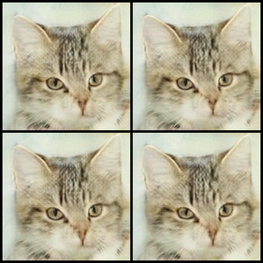}
  \end{subfigure}
  \begin{subfigure}[t]{0.19\textwidth}
    \centering
    \includegraphics[width=\linewidth]{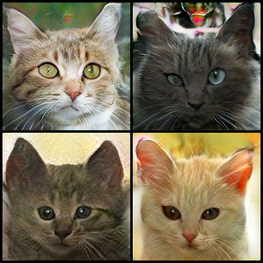}
  \end{subfigure}
  \begin{subfigure}[t]{0.19\textwidth}
    \centering
    \includegraphics[width=\linewidth]{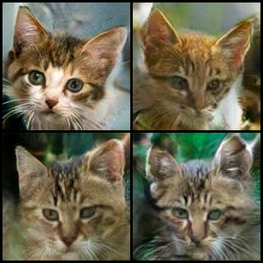}
  \end{subfigure}

  \begin{subfigure}[t]{0.19\textwidth}
    \centering
    \includegraphics[width=\linewidth]{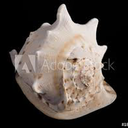}
    \caption{Shells}
    \label{fig-visual-recall:shells}
  \end{subfigure}
  \begin{subfigure}[t]{0.19\textwidth}
    \centering
    \includegraphics[width=\linewidth]{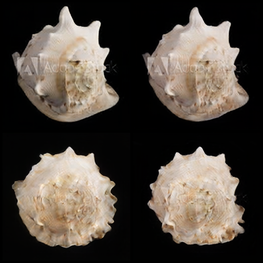}
  \end{subfigure}
  \begin{subfigure}[t]{0.19\textwidth}
    \centering
    \includegraphics[width=\linewidth]{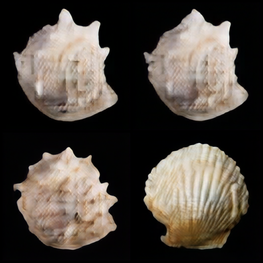}
  \end{subfigure}
  \begin{subfigure}[t]{0.19\textwidth}
    \centering
    \includegraphics[width=\linewidth]{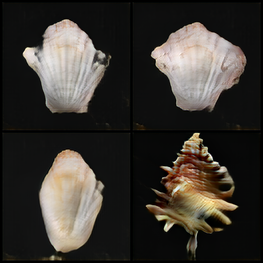}
  \end{subfigure}
  \begin{subfigure}[t]{0.19\textwidth}
    \centering
    \includegraphics[width=\linewidth]{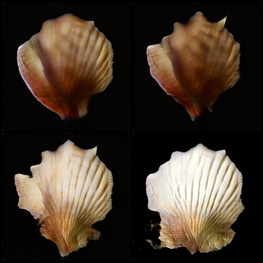}
  \end{subfigure}

  \begin{subfigure}[t]{0.19\textwidth}
    \centering
    \includegraphics[width=\linewidth]{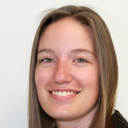}
    \caption{FFHQ-100}
    \label{fig-visual-recall:ffhq}
  \end{subfigure}
  \begin{subfigure}[t]{0.19\textwidth}
    \centering
    \includegraphics[width=\linewidth]{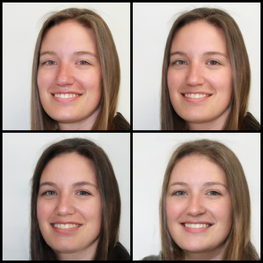}
  \end{subfigure}
  \begin{subfigure}[t]{0.19\textwidth}
    \centering
    \includegraphics[width=\linewidth]{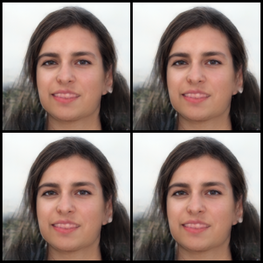}
  \end{subfigure}
  \begin{subfigure}[t]{0.19\textwidth}
    \centering
    \includegraphics[width=\linewidth]{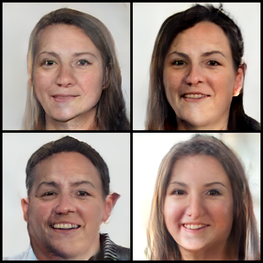}
  \end{subfigure}
  \begin{subfigure}[t]{0.19\textwidth}
    \centering
    \includegraphics[width=\linewidth]{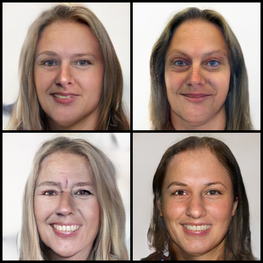}
  \end{subfigure}

  \caption{\textbf{Visual Recall test.} The first column is the query image from the dataset. Subsequent columns are the samples produced by different methods that are closest to the query image in LPIPS feature space. The samples produced by our method are closer to the query images compared to the baselines, while remaining diverse.}
  \label{fig-visual-recall}
\end{figure*}

\subsection{Qualitative Results}

Figure \ref{fig-random-samples} compares the random samples of our method to that of several baselines, and we observe that our method produces overall sharper and more diverse images. In addition, we propose Visual Recall, a simple test to substantiate the qualitative superiority of our method. For each method, we first generate 1000 samples. Next, using a real image from the dataset as a query, we find the images from the pool of generated samples that are closest to the query image. We use LPIPS features \cite{lpips} for the computing the distance between real images and samples. Figure~\ref{fig-visual-recall} shows the results for the proposed test for different datasets and methods. 
We see that the samples produced by our method are visually similar to the query, while being sharp and diverse in attributes like hair colour, smile and jaw structure. Note that other methods do not have samples that closely resemble the query image.

\begin{figure*}[!t]
  \captionsetup[subfigure]{labelformat=parens}

  \begin{subfigure}[t]{\textwidth}
    \centering
    \includegraphics[width=\linewidth]{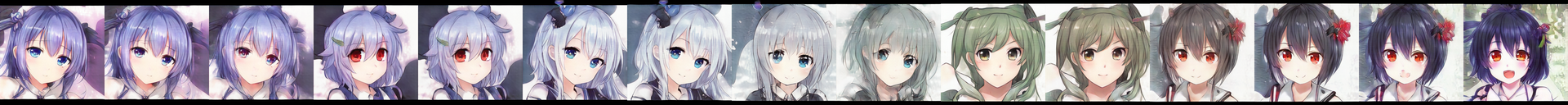}
    \includegraphics[width=\linewidth]{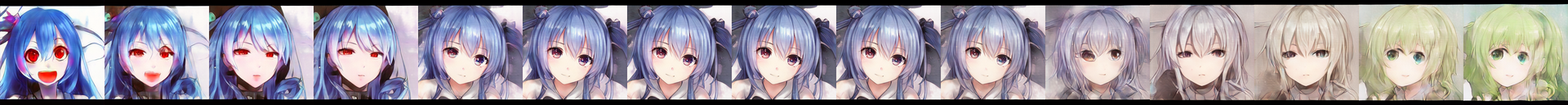}
    \includegraphics[width=\linewidth]{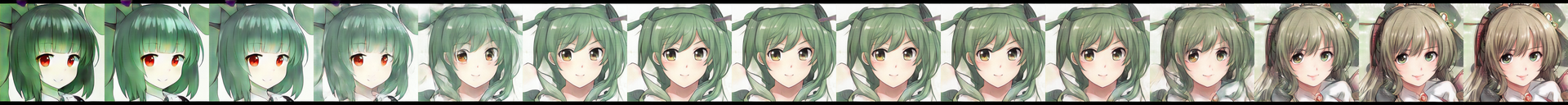}
    \caption{{Anime}}
    \label{interpolate-anime}
  \end{subfigure}

  \begin{subfigure}[t]{\textwidth}
    \centering
    \includegraphics[width=\linewidth]{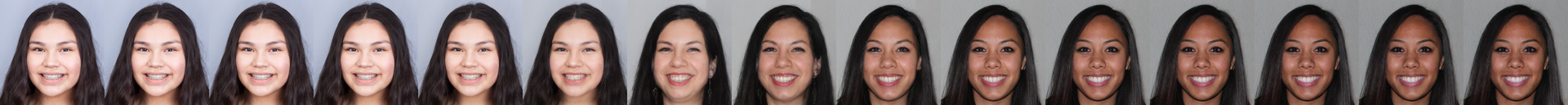}
    \includegraphics[width=\linewidth]{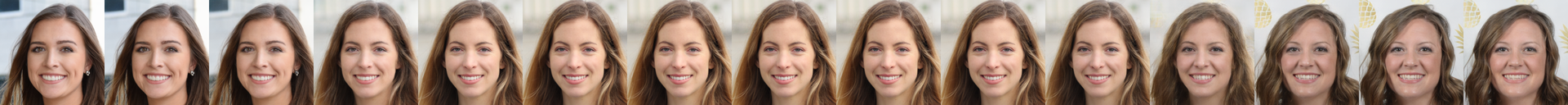}
    \includegraphics[width=\linewidth]{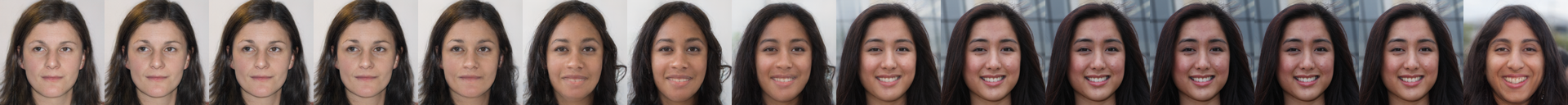}
    \caption{{FFHQ-100}}
    \label{interpolate-ffhq}
  \end{subfigure}

  \begin{subfigure}[t]{\textwidth}
    \centering
    \includegraphics[width=\linewidth]{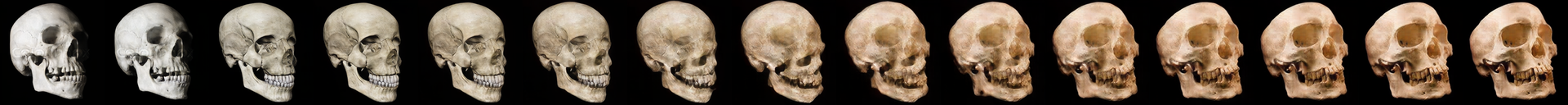}
    \includegraphics[width=\linewidth]{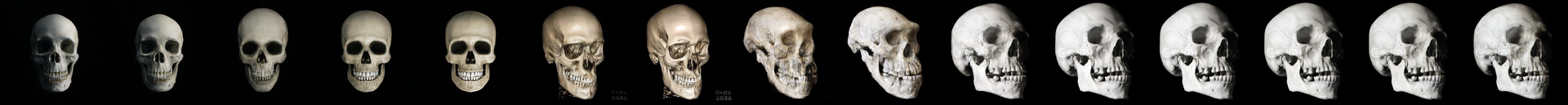}
    \includegraphics[width=\linewidth]{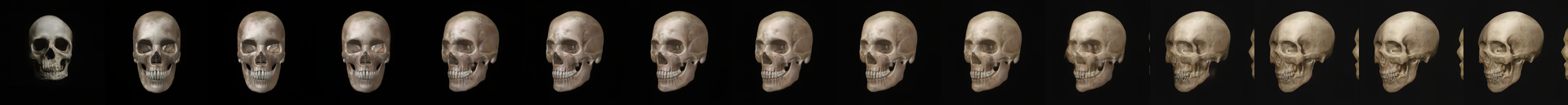}
    \caption{{Skulls}}
    \label{interpolate-skulls}
  \end{subfigure}

  \caption{\textbf{Latent space interpolation}. We observe that the output images changes smoothly in a meaningful manner.}
  \label{fig-latent-interpolate}
\end{figure*}

Figure~\ref{fig-latent-interpolate} shows results of spherical linear interpolation between two random points in the latent space for different datasets. The images transition in a meaningful manner, indicating that our model has learnt a continuous and structured latent space representation of the image distribution.

\subsection{Ablation Study}

Table-\ref{tab:eps-ablation} presents the FID computed for different values of $\epsilon$ for three datasets. We observe that our approach works best for values of $\epsilon$ close to 0.15 and that increasing the value of $\epsilon$ beyond a certain range degrades the performance. 

\begin{table}[]
\centering
\footnotesize
\begin{tabularx}{0.7\linewidth}{nnnnn}
\toprule
\textbf{Dataset} & 0.12 & 0.15 & 0.18 & 0.22 \\ 
\midrule
FFHQ-100 & 13.6  & \textbf{12.91} & 13.01 & 14.8  \\ 
Obama    & 14.44 & \textbf{14.03} & 14.62 & 14.34 \\ 
Cat      & 17.29 & \textbf{15.96} & 16.12 & 16.27 \\ 
\bottomrule
\end{tabularx}%
\caption{FID for different values of $\epsilon$}
\label{tab:eps-ablation}
\end{table}

\section{Conclusion}
In this paper, we identified a latent space misalignment between the training and testing phases of existing IMLE-based methods, resulting in poor performance in few-shot image synthesis tasks. To address this issue, we introduced a novel algorithm, RS-IMLE, which modifies the prior distribution used for training. Our experimental results demonstrate that our method significantly enhances the quality of generated images and mode coverage during inference. ~\\ \\
\textbf{Acknowledgements: } 
This research was enabled in part by support provided by NSERC, the BC DRI Group and the Digital Research Alliance of Canada. The authors would also like to thank Tristan Engst for extensive help polishing our paper.

\clearpage

\bibliographystyle{splncs04}
\bibliography{egbib}

\begin{thebibliography}{10}
\providecommand{\url}[1]{\texttt{#1}}
\providecommand{\urlprefix}{URL }
\providecommand{\doi}[1]{https://doi.org/#1}

\bibitem{adaptiveIMLE}
Aghabozorgi, M., Peng, S., Li, K.: Adaptive {IMLE} for few-shot pretraining-free generative modelling. In: Krause, A., Brunskill, E., Cho, K., Engelhardt, B., Sabato, S., Scarlett, J. (eds.) Proceedings of the 40th International Conference on Machine Learning. Proceedings of Machine Learning Research, vol.~202, pp. 248--264. PMLR (23--29 Jul 2023), \url{https://proceedings.mlr.press/v202/aghabozorgi23a.html}

\bibitem{biggan}
Brock, A., Donahue, J., Simonyan, K.: Large scale gan training for high fidelity natural image synthesis. ArXiv  \textbf{abs/1809.11096} (2019)

\bibitem{vdvae}
Child, R.: Very deep vaes generalize autoregressive models and can outperform them on images. ArXiv  \textbf{abs/2011.10650} (2021)

\bibitem{Dhariwal2021DiffusionMB}
Dhariwal, P., Nichol, A.: Diffusion models beat gans on image synthesis. ArXiv  \textbf{abs/2105.05233} (2021)

\bibitem{Dinh2017DensityEU}
Dinh, L., Sohl-Dickstein, J.N., Bengio, S.: Density estimation using real nvp. ArXiv  \textbf{abs/1605.08803} (2017)

\bibitem{Esser2020TamingTF}
Esser, P., Rombach, R., Ommer, B.: Taming transformers for high-resolution image synthesis. 2021 IEEE/CVF Conference on Computer Vision and Pattern Recognition (CVPR) pp. 12868--12878 (2020), \url{https://api.semanticscholar.org/CorpusID:229297973}

\bibitem{goodfellow2014generative}
Goodfellow, I., Pouget-Abadie, J., Mirza, M., Xu, B., Warde-Farley, D., Ozair, S., Courville, A., Bengio, Y.: Generative adversarial nets. Advances in neural information processing systems  \textbf{27} (2014)

\bibitem{fid}
Heusel, M., Ramsauer, H., Unterthiner, T., Nessler, B., Hochreiter, S.: Gans trained by a two time-scale update rule converge to a local nash equilibrium. In: NIPS (2017)

\bibitem{Ho2020DenoisingDP}
Ho, J., Jain, A., Abbeel, P.: Denoising diffusion probabilistic models. ArXiv  \textbf{abs/2006.11239} (2020)

\bibitem{stylegan3}
Karras, T., Aittala, M., Laine, S., H{\"a}rk{\"o}nen, E., Hellsten, J., Lehtinen, J., Aila, T.: Alias-free generative adversarial networks. In: NeurIPS (2021)

\bibitem{karras2019style}
Karras, T., Laine, S., Aila, T.: A style-based generator architecture for generative adversarial networks. In: Proceedings of the IEEE/CVF conference on computer vision and pattern recognition. pp. 4401--4410 (2019)

\bibitem{karras2019stylebased}
Karras, T., Laine, S., Aila, T.: A style-based generator architecture for generative adversarial networks (2019)

\bibitem{Karras2019AnalyzingAI}
Karras, T., Laine, S., Aittala, M., Hellsten, J., Lehtinen, J., Aila, T.: Analyzing and improving the image quality of stylegan. 2020 IEEE/CVF Conference on Computer Vision and Pattern Recognition (CVPR) pp. 8107--8116 (2019), \url{https://api.semanticscholar.org/CorpusID:209202273}

\bibitem{stylegan2}
Karras, T., Laine, S., Aittala, M., Hellsten, J., Lehtinen, J., Aila, T.: Analyzing and improving the image quality of stylegan. 2020 IEEE/CVF Conference on Computer Vision and Pattern Recognition (CVPR) pp. 8107--8116 (2020)

\bibitem{Kingma2018GlowGF}
Kingma, D.P., Dhariwal, P.: Glow: Generative flow with invertible 1x1 convolutions. ArXiv  \textbf{abs/1807.03039} (2018)

\bibitem{kingma2013auto}
Kingma, D.P., Welling, M.: Auto-encoding variational bayes. arXiv preprint arXiv:1312.6114  (2013)

\bibitem{Kobyzev2021NormalizingFA}
Kobyzev, I., Prince, S., Brubaker, M.A.: Normalizing flows: An introduction and review of current methods. IEEE Transactions on Pattern Analysis and Machine Intelligence  \textbf{43},  3964--3979 (2021)

\bibitem{mixdl}
Kong, C., Kim, J., Han, D., Kwak, N.: Few-shot image generation with mixup-based distance learning (2022)

\bibitem{kynkaanniemi2019improved}
Kynk{\"a}{\"a}nniemi, T., Karras, T., Laine, S., Lehtinen, J., Aila, T.: Improved precision and recall metric for assessing generative models. Advances in Neural Information Processing Systems  \textbf{32} (2019)

\bibitem{li2017fast}
Li, K., Malik, J.: Fast k-nearest neighbour search via prioritized dci (2017)

\bibitem{li2018implicit}
Li, K., Malik, J.: Implicit maximum likelihood estimation. arXiv preprint arXiv:1809.09087  (2018)

\bibitem{Li2020FewshotIG}
Li, Y., Zhang, R., Lu, J., Shechtman, E.: Few-shot image generation with elastic weight consolidation. ArXiv  \textbf{abs/2012.02780} (2020)

\bibitem{fakeclr}
Li, Z., Wang, C., Zheng, H., Zhang, J., Li, B.: Fakeclr: Exploring contrastive learning for solving latent discontinuity in data-efficient gans. In: ECCV (2022)

\bibitem{DBLP:fastgan}
Liu, B., Zhu, Y., Song, K., Elgammal, A.: Towards faster and stabilized {GAN} training for high-fidelity few-shot image synthesis. CoRR  \textbf{abs/2101.04775} (2021), \url{https://arxiv.org/abs/2101.04775}

\bibitem{Mo2020FreezeDA}
Mo, S., Cho, M., Shin, J.: Freeze discriminator: A simple baseline for fine-tuning gans. ArXiv  \textbf{abs/2002.10964} (2020)

\bibitem{Ojha2021FewshotIG}
Ojha, U., Li, Y., Lu, J., Efros, A.A., Lee, Y.J., Shechtman, E., Zhang, R.: Few-shot image generation via cross-domain correspondence. 2021 IEEE/CVF Conference on Computer Vision and Pattern Recognition (CVPR) pp. 10738--10747 (2021)

\bibitem{Oord2016ConditionalIG}
van~den Oord, A., Kalchbrenner, N., Espeholt, L., Kavukcuoglu, K., Vinyals, O., Graves, A.: Conditional image generation with pixelcnn decoders. ArXiv  \textbf{abs/1606.05328} (2016)

\bibitem{Oord2016PixelRN}
van~den Oord, A., Kalchbrenner, N., Kavukcuoglu, K.: Pixel recurrent neural networks. ArXiv  \textbf{abs/1601.06759} (2016)

\bibitem{Razavi2019GeneratingDH}
Razavi, A., van~den Oord, A., Vinyals, O.: Generating diverse high-fidelity images with vq-vae-2. ArXiv  \textbf{abs/1906.00446} (2019)

\bibitem{Salimans2017PixelCNNIT}
Salimans, T., Karpathy, A., Chen, X., Kingma, D.P.: Pixelcnn++: Improving the pixelcnn with discretized logistic mixture likelihood and other modifications. ArXiv  \textbf{abs/1701.05517} (2017)

\bibitem{Sauer2021ProjectedGC}
Sauer, A., Chitta, K., Muller, J., Geiger, A.: Projected gans converge faster. In: Neural Information Processing Systems (2021), \url{https://api.semanticscholar.org/CorpusID:240354401}

\bibitem{ReGAN}
Saxena, D., Cao, J., Xu, J., Kulshrestha, T.: Re-gan: Data-efficient gans training via architectural reconfiguration. In: Proceedings of the IEEE/CVF Conference on Computer Vision and Pattern Recognition (CVPR). pp. 16230--16240 (June 2023)

\bibitem{dataset-cg}
Si, Z., Zhu, S.C.: Learning hybrid image templates (hit) by information projection. IEEE Transactions on Pattern Analysis and Machine Intelligence  \textbf{34},  1354--1367 (2012)

\bibitem{Song2019GenerativeMB}
Song, Y., Ermon, S.: Generative modeling by estimating gradients of the data distribution. ArXiv  \textbf{abs/1907.05600} (2019)

\bibitem{Song2021ScoreBasedGM}
Song, Y., Sohl-Dickstein, J.N., Kingma, D.P., Kumar, A., Ermon, S., Poole, B.: Score-based generative modeling through stochastic differential equations. ArXiv  \textbf{abs/2011.13456} (2021)

\bibitem{Vahdat2020NVAEAD}
Vahdat, A., Kautz, J.: Nvae: A deep hierarchical variational autoencoder. ArXiv  \textbf{abs/2007.03898} (2020)

\bibitem{Wang2020MineGANEK}
Wang, Y., Gonzalez-Garcia, A., Berga, D., Herranz, L., Khan, F.S., van~de Weijer, J.: Minegan: Effective knowledge transfer from gans to target domains with few images. 2020 IEEE/CVF Conference on Computer Vision and Pattern Recognition (CVPR) pp. 9329--9338 (2020)

\bibitem{yang2022FreGAN}
Yang, M., Wang, Z., Chi, Z., Zhang, Y.: Fregan: Exploiting frequency components for training gans under limited data  (2022)

\bibitem{lpips}
Zhang, R., Isola, P., Efros, A.A., Shechtman, E., Wang, O.: The unreasonable effectiveness of deep features as a perceptual metric. In: CVPR (2018)

\bibitem{Zhao2020OnLP}
Zhao, M., Cong, Y., Carin, L.: On leveraging pretrained gans for generation with limited data. In: ICML (2020)

\bibitem{Zhao2020DifferentiableAF}
Zhao, S., Liu, Z., Lin, J., Zhu, J.Y., Han, S.: Differentiable augmentation for data-efficient gan training. ArXiv  \textbf{abs/2006.10738} (2020)

\bibitem{precrecall}
Zhu, P., Abdal, R., Qin, Y., Wonka, P.: Improved stylegan embedding: Where are the good latents? ArXiv  \textbf{abs/2012.09036} (2020)

\end{thebibliography}
\clearpage

\appendix 

\section{Theory}

Since we use rejection sampling, we need to ensure that the acceptance ratio is bounded. Recall that: 

\begin{align}
    & m \left(1 - F_{\Tilde{D}_{i1}}(t)\right)^{m-1} f_{\Tilde{D}_{i1}}(t) = n \left(1 - F_{{D}_{i1}}(t)\right)^{n-1} f_{{D}_{i1}}(t) \nonumber \\
    &\implies f_{\Tilde{D}_{i1}}(t) = \frac{n}{m} \frac{\left(1 - F_{{D}_{i1}}(t)\right)^{n-1}}{\left(1 - F_{\Tilde{D}_{i1}}(t)\right)^{m-1}} f_{{D}_{i1}}(t) \label{eqn:f-tilde}
\end{align}

Notice that because of $\left(1 - F_{\Tilde{D}_{i1}}(t)\right)^{m-1}$ term in the denominator, we have to make sure that the expression for $f_{\Tilde{D}_{i1}}(t)$ is bounded. One way to do that is to truncate the right tail of the ideal distribution $f_{{D}_{i1}}(t)$ to $0$. More explicitly, for a very large $T$ (e.g., $T = 100,000$), we can write:

\begin{align*}
    g_{{D}_{i1}}(t) = 
    \begin{cases} 
    f_{{D}_{i1}}(t) & \text{if } t \leq T \\
    0 & \text{if } t > T
    \end{cases}
\end{align*}

Here $g_{{D}_{i1}}(t)$ is the PDF of the truncated distribution. Since very large values of distances ($t$) are rarely observed at test time, so applying this truncation has little effect in practice. Instead of writing the expression for Equation \ref{eqn:f-tilde} in terms of $g_{{D}_{i1}}(t)$, we continue to use $f_{{D}_{i1}}(t)$ along with a constant $c$ associated with the truncation.

Hence using $\phi(t) = \frac{n}{m} \frac{\left(1 - F_{{D}_{i1}}(t)\right)^{n-1}}{\left(1 - F_{\Tilde{D}_{i1}}(t)\right)^{m-1}} $ and $c$ as the constant associated with the truncation described above, we can write Equation \ref{eqn:f-tilde} as:

\begin{align}
    f_{\Tilde{D}_{i1}}(t) = c \phi(t) f_{{D}_{i1}}(t)
\end{align}

\section{Network Architecture}
\begin{figure}[ht]
    \centering
    \subfloat[Network Architecture]{
    \includegraphics[width=0.53\linewidth]{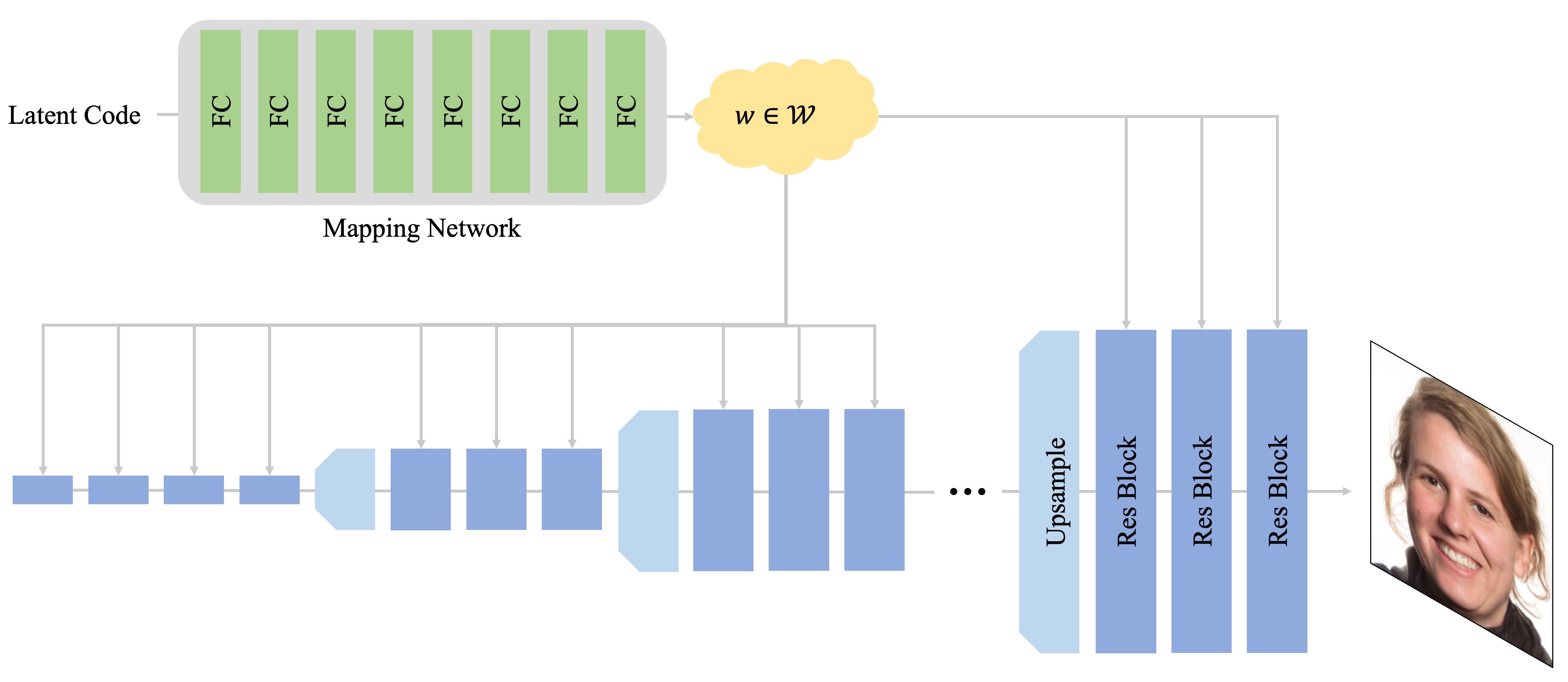}
    \label{fig:overall}
  }
  \subfloat[Res Block]{
    \includegraphics[width=0.27\linewidth]{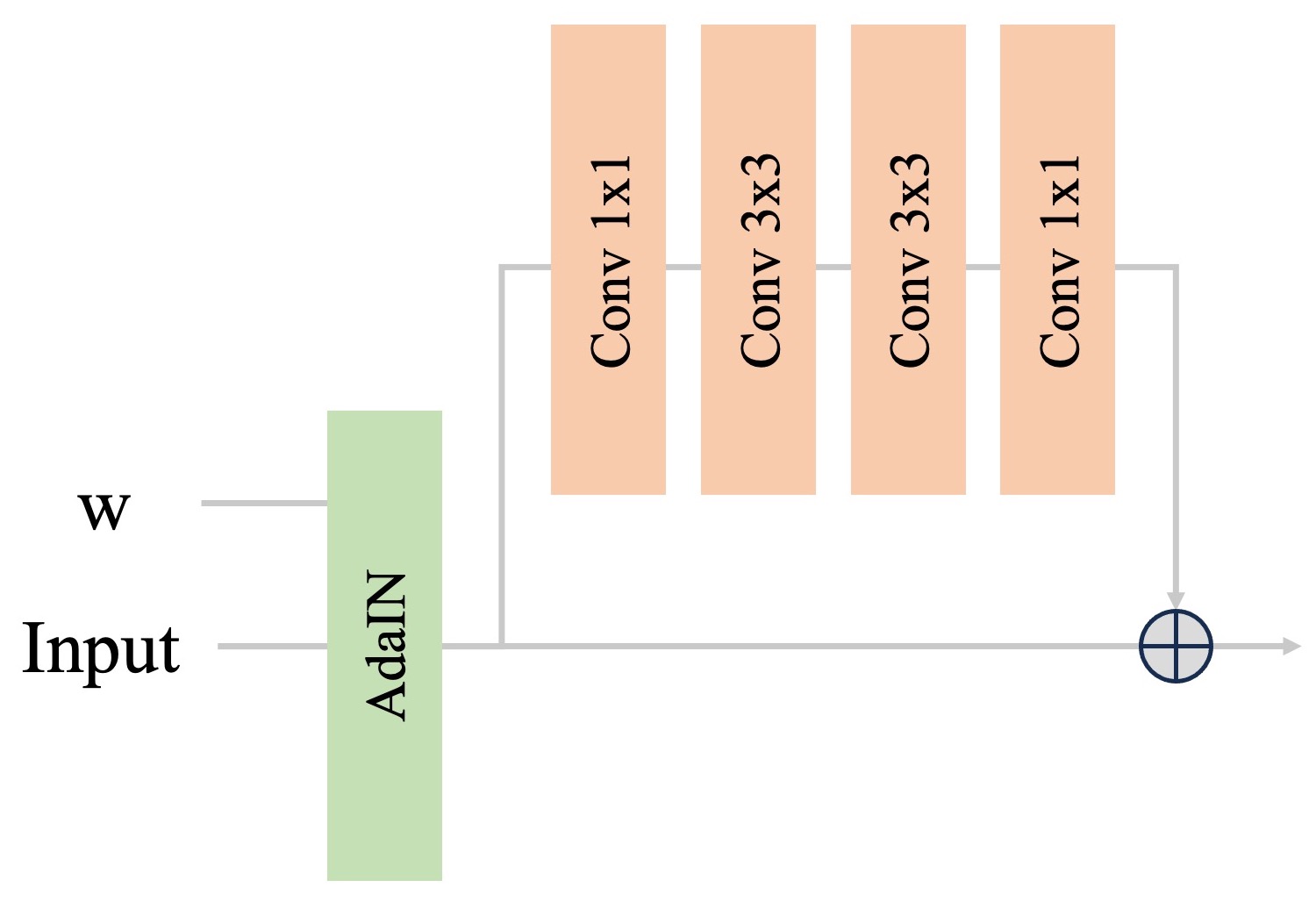}
    \label{fig:res}
  }
    \caption{(a) Network architecture, which comprises of a mapping network, upsampling layers and res blocks (details in (b)). (b) Inner workings of res blocks. }
    \label{fig:arch}
\end{figure}

Our network architecture is illustrated in Figure~\ref{fig:arch}, comprising a fully-connected mapping network inspired by \cite{karras2019style} and a generator network constructed using decoder modules from VDVAE~\cite{vdvae}. We choose an input latent dimension of $1024$ for all datasets.

\section{Experiments}

Table-\ref{tab:dataset-info} gives the details about the number of images in each dataset as well as the value of radius used in the rejection sampling procedure (epsilon, $\epsilon$) used in the results presented in the main paper. The selection of epsilon values was conducted through the process of hyperparameter tuning. We present an ablation study with different values of epsilon later in the paper.

\begin{table}[]
\centering
\begin{tabularx}{\linewidth}{lnnnnnnnnn}
\toprule
 & Obama & Grumpy Cat & Panda & FFHQ-100 & Cat  & Dog  & Anime & Skulls & Shells \\ 
\midrule
Num. of Images & 100   & 100        & 100   & 100      & 160  & 389  & 120   & 96     & 64  \\   
Epsilon Used    & 0.15  & 0.18       & 0.18  & 0.15     & 0.15 & 0.15 & 0.18  & 0.18   & 0.18   \\ 
\bottomrule
\end{tabularx}%
\caption{Number of images in each dataset and the value of epsilon used.}
\label{tab:dataset-info}
\end{table}

\subsection{Random samples}
\begin{figure*}[]
  \captionsetup[subfigure]{labelformat=parens}  
  \begin{subfigure}[t]{0.16\textwidth}
    \centering
    FastGAN
  \end{subfigure}
  \begin{subfigure}[t]{0.16\textwidth}
    \centering
    FakeCLR
  \end{subfigure}
  \begin{subfigure}[t]{0.16\textwidth}
    \centering
    FreGAN
  \end{subfigure}
  \begin{subfigure}[t]{0.16\textwidth}
    \centering
    ReGAN
  \end{subfigure}
  \begin{subfigure}[t]{0.16\textwidth}
    \centering
    AdaIMLE
  \end{subfigure}
  \begin{subfigure}[t]{0.16\textwidth}
    \centering
    \textit{Ours}
  \end{subfigure}
  
  \vspace{1em}

  \begin{subfigure}[t]{0.16\textwidth}
    \centering
    \includegraphics[width=\linewidth]{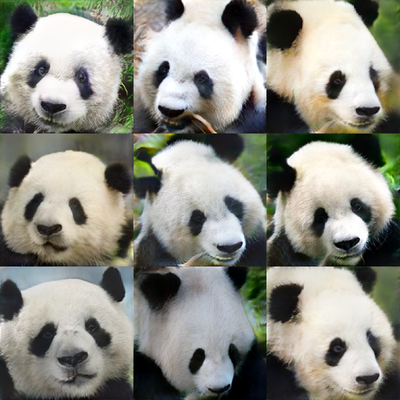}
  \end{subfigure}
  \begin{subfigure}[t]{0.16\textwidth}
    \centering
    \includegraphics[width=\linewidth]{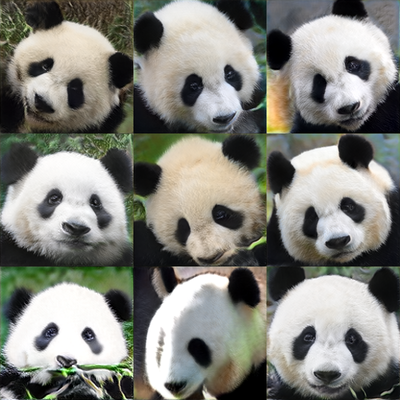}
  \end{subfigure}
  \begin{subfigure}[t]{0.16\textwidth}
    \centering
    \includegraphics[width=\linewidth]{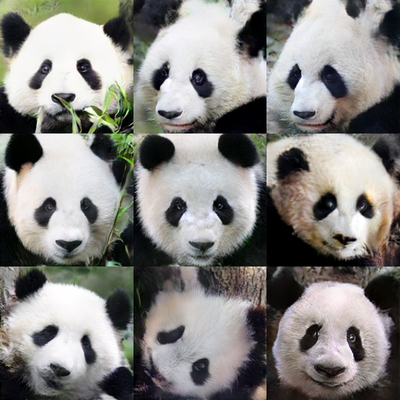}
  \end{subfigure}
  \begin{subfigure}[t]{0.16\textwidth}
    \centering
    \includegraphics[width=\linewidth]{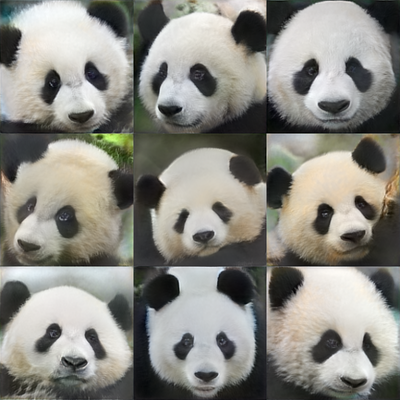}
  \end{subfigure}
  \begin{subfigure}[t]{0.16\textwidth}
    \centering
    \includegraphics[width=\linewidth]{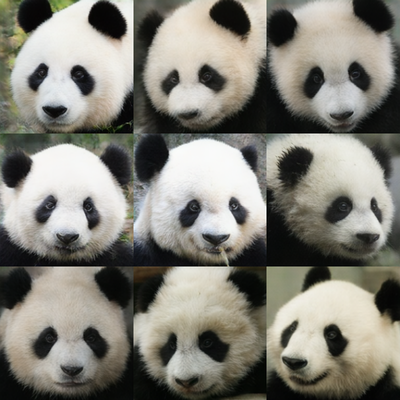}
  \end{subfigure}
  \begin{subfigure}[t]{0.16\textwidth}
    \centering
    \includegraphics[width=\linewidth]{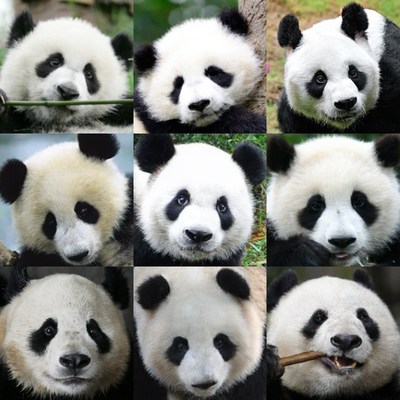}
  \end{subfigure}

  \vspace{0.5em}

  \begin{subfigure}[t]{0.16\textwidth}
    \centering
    \includegraphics[width=\linewidth]{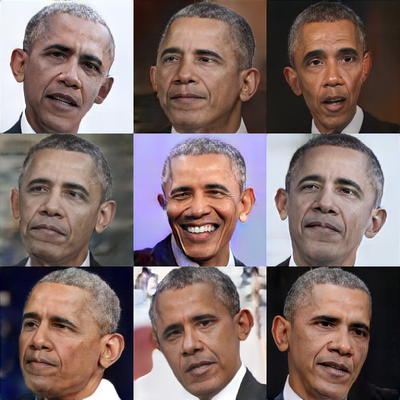}
  \end{subfigure}
  \begin{subfigure}[t]{0.16\textwidth}
    \centering
    \includegraphics[width=\linewidth]{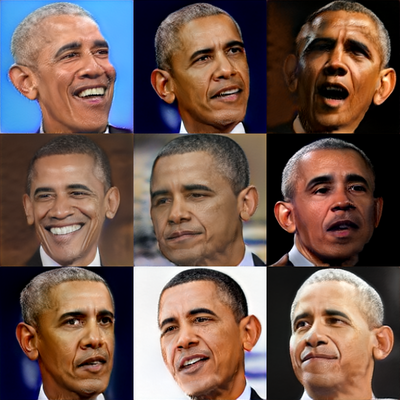}
  \end{subfigure}
  \begin{subfigure}[t]{0.16\textwidth}
    \centering
    \includegraphics[width=\linewidth]{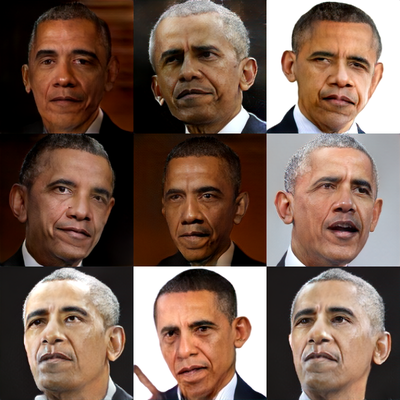}
  \end{subfigure}
  \begin{subfigure}[t]{0.16\textwidth}
    \centering
    \includegraphics[width=\linewidth]{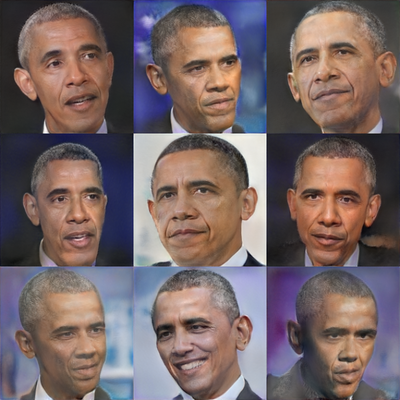}
  \end{subfigure}
  \begin{subfigure}[t]{0.16\textwidth}
    \centering
    \includegraphics[width=\linewidth]{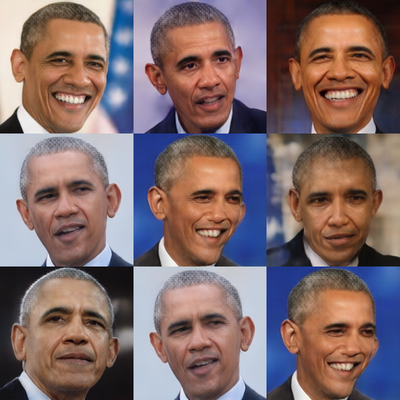}
  \end{subfigure}
  \begin{subfigure}[t]{0.16\textwidth}
    \centering
    \includegraphics[width=\linewidth]{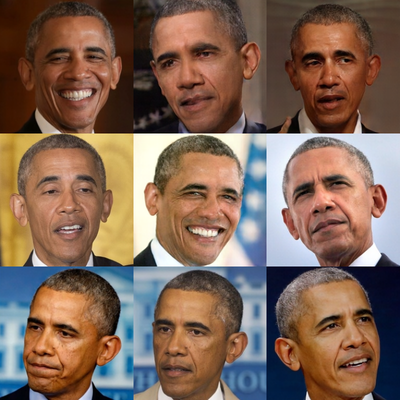}
  \end{subfigure}

  \vspace{0.5em}
  
  \begin{subfigure}[t]{0.16\textwidth}
    \centering
    \includegraphics[width=\linewidth]{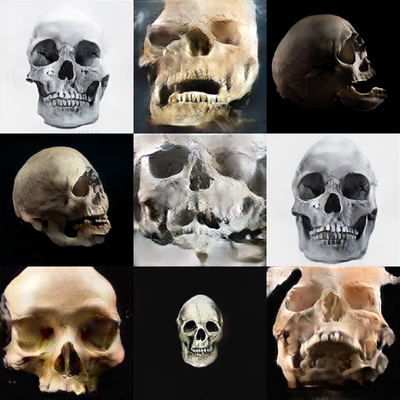}
  \end{subfigure}
  \begin{subfigure}[t]{0.16\textwidth}
    \centering
    \includegraphics[width=\linewidth]{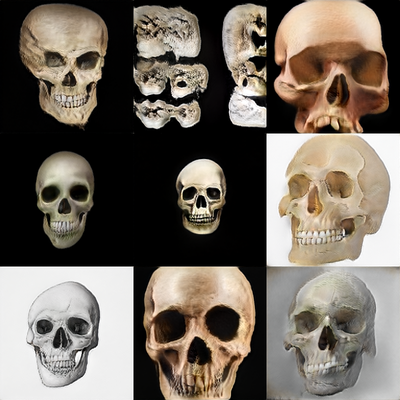}
  \end{subfigure}
  \begin{subfigure}[t]{0.16\textwidth}
    \centering
    \includegraphics[width=\linewidth]{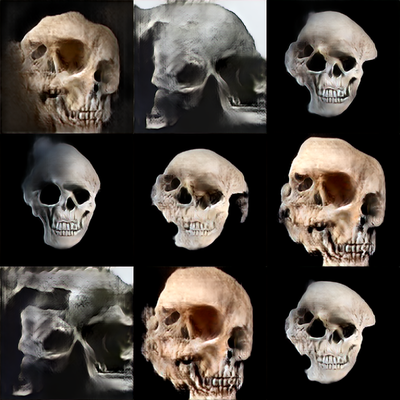}
  \end{subfigure}
  \begin{subfigure}[t]{0.16\textwidth}
    \centering
    \includegraphics[width=\linewidth]{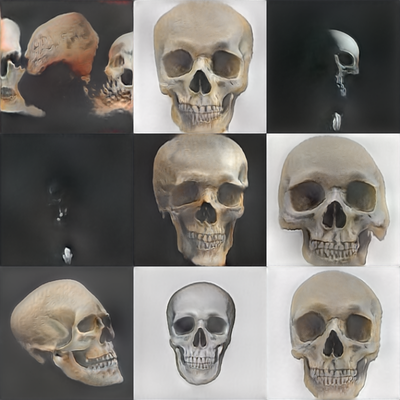}
  \end{subfigure}
  \begin{subfigure}[t]{0.16\textwidth}
    \centering
    \includegraphics[width=\linewidth]{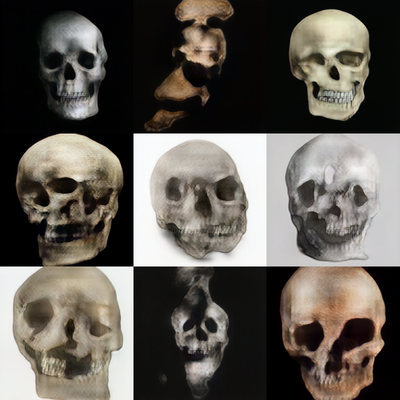}
  \end{subfigure}
  \begin{subfigure}[t]{0.16\textwidth}
    \centering
    \includegraphics[width=\linewidth]{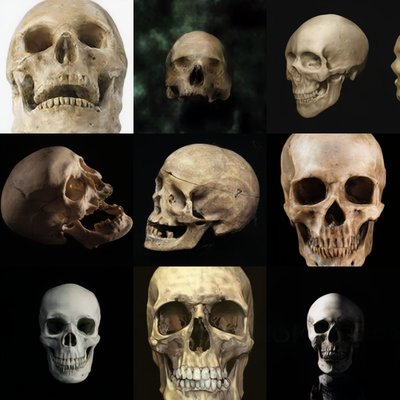}
    \end{subfigure}

  \vspace{0.5em}

    \begin{subfigure}[t]{0.16\textwidth}
    \centering
    \includegraphics[width=\linewidth]{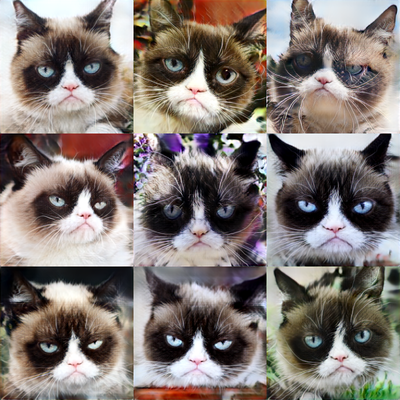}
  \end{subfigure}
  \begin{subfigure}[t]{0.16\textwidth}
    \centering
    \includegraphics[width=\linewidth]{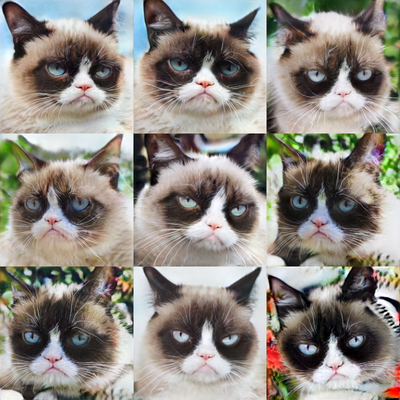}
  \end{subfigure}
  \begin{subfigure}[t]{0.16\textwidth}
    \centering
    \includegraphics[width=\linewidth]{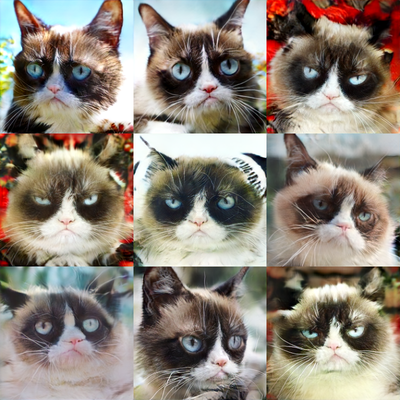}
  \end{subfigure}
  \begin{subfigure}[t]{0.16\textwidth}
    \centering
    \includegraphics[width=\linewidth]{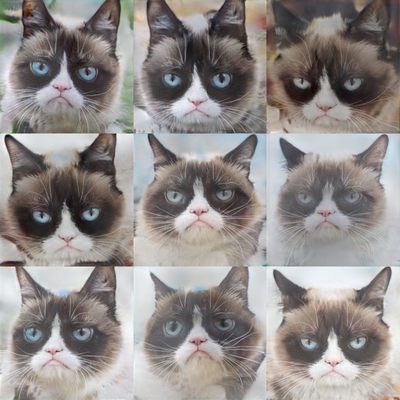}
  \end{subfigure}
  \begin{subfigure}[t]{0.16\textwidth}
    \centering
    \includegraphics[width=\linewidth]{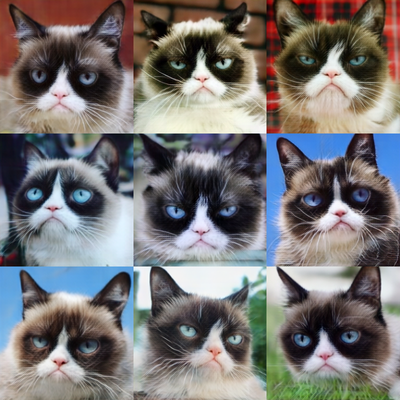}
  \end{subfigure}
  \begin{subfigure}[t]{0.16\textwidth}
    \centering
    \includegraphics[width=\linewidth]{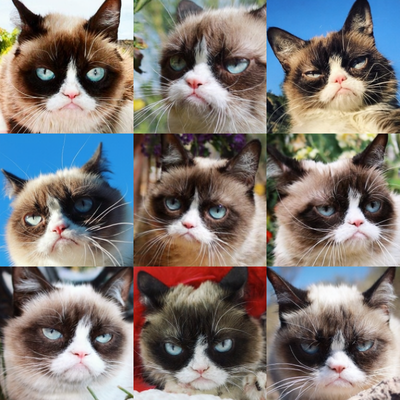}
  \end{subfigure}

  \vspace{0.5em}

  \begin{subfigure}[t]{0.16\textwidth}
    \centering
    \includegraphics[width=\linewidth]{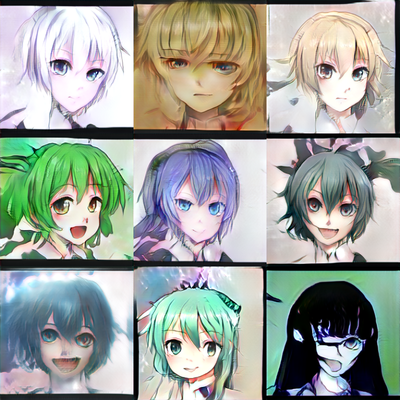}
  \end{subfigure}
  \begin{subfigure}[t]{0.16\textwidth}
    \centering
    \includegraphics[width=\linewidth]{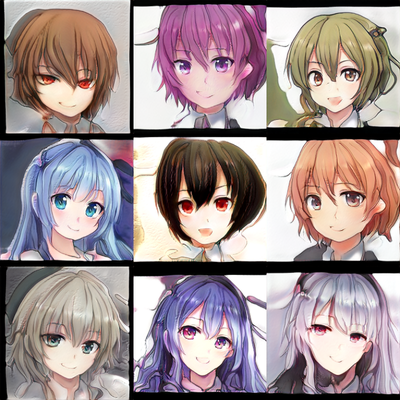}
  \end{subfigure}
  \begin{subfigure}[t]{0.16\textwidth}
    \centering
    \includegraphics[width=\linewidth]{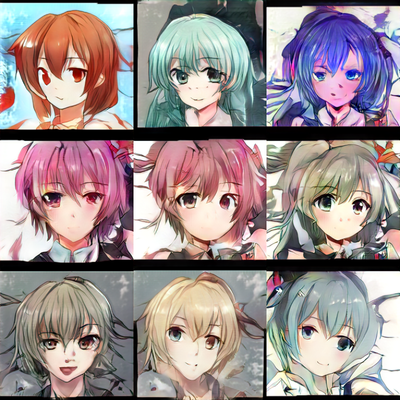}
  \end{subfigure}
  \begin{subfigure}[t]{0.16\textwidth}
    \centering
    \includegraphics[width=\linewidth]{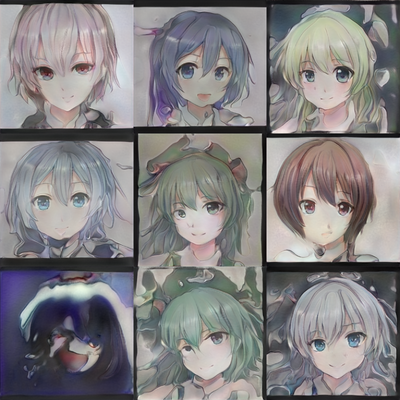}
  \end{subfigure}
  \begin{subfigure}[t]{0.16\textwidth}
    \centering
    \includegraphics[width=\linewidth]{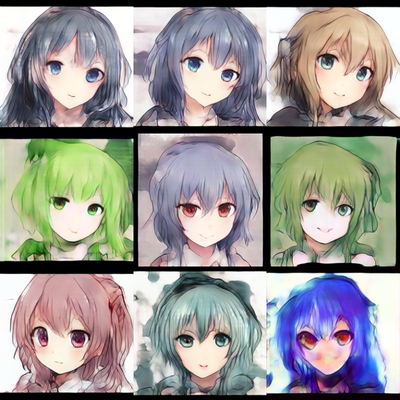}
  \end{subfigure}
  \begin{subfigure}[t]{0.16\textwidth}
    \centering
    \includegraphics[width=\linewidth]{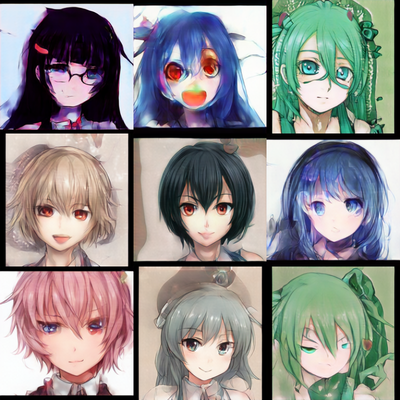}
  \end{subfigure}

  \vspace{0.5em}

  \begin{subfigure}[t]{0.16\textwidth}
    \centering
    \includegraphics[width=\linewidth]{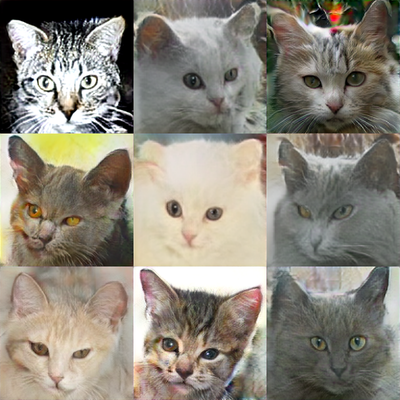}
  \end{subfigure}
  \begin{subfigure}[t]{0.16\textwidth}
    \centering
    \includegraphics[width=\linewidth]{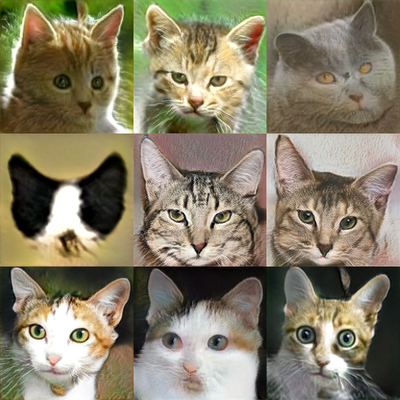}
  \end{subfigure}
  \begin{subfigure}[t]{0.16\textwidth}
    \centering
    \includegraphics[width=\linewidth]{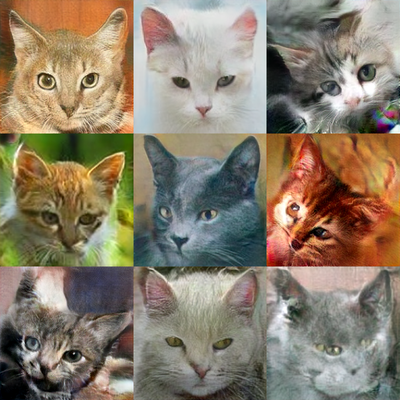}
  \end{subfigure}
  \begin{subfigure}[t]{0.16\textwidth}
    \centering
    \includegraphics[width=\linewidth]{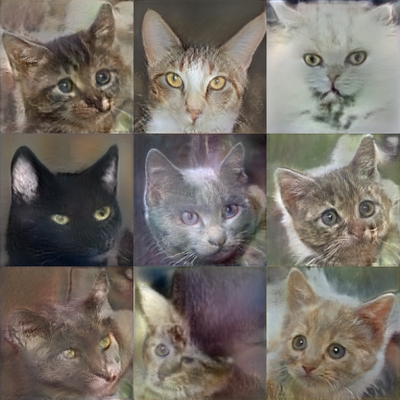}
  \end{subfigure}
  \begin{subfigure}[t]{0.16\textwidth}
    \centering
    \includegraphics[width=\linewidth]{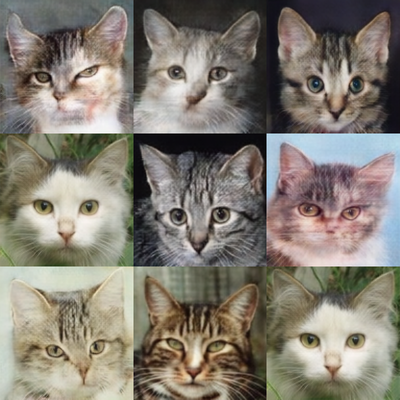}
  \end{subfigure}
  \begin{subfigure}[t]{0.16\textwidth}
    \centering
    \includegraphics[width=\linewidth]{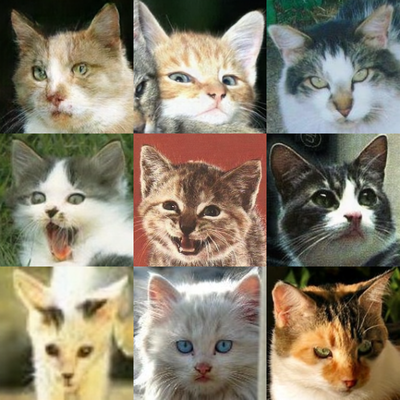}
  \end{subfigure}

  \vspace{0.5em}

  \begin{subfigure}[t]{0.16\textwidth}
    \centering
    \includegraphics[width=\linewidth]{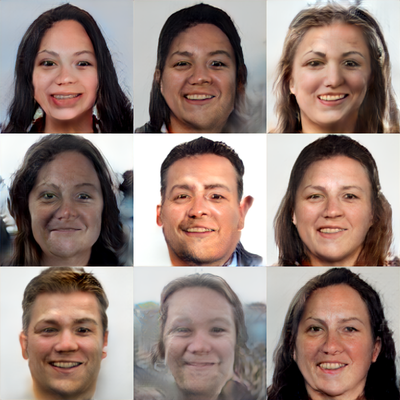}
  \end{subfigure}
  \begin{subfigure}[t]{0.16\textwidth}
    \centering
    \includegraphics[width=\linewidth]{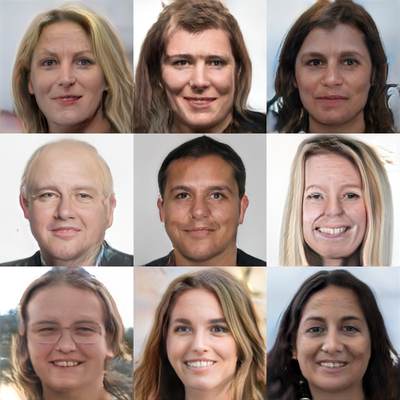}
  \end{subfigure}
  \begin{subfigure}[t]{0.16\textwidth}
    \centering
    \includegraphics[width=\linewidth]{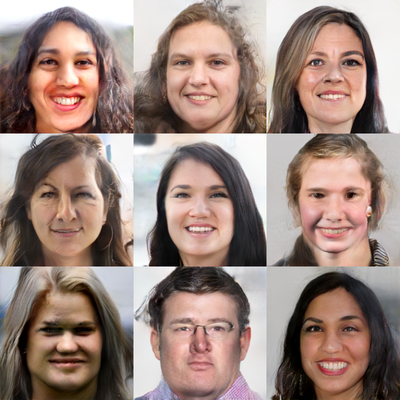}
  \end{subfigure}
  \begin{subfigure}[t]{0.16\textwidth}
    \centering
    \includegraphics[width=\linewidth]{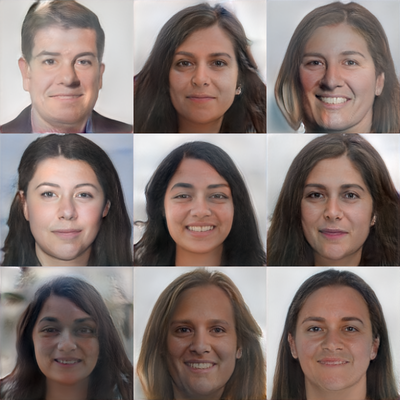}
  \end{subfigure}
  \begin{subfigure}[t]{0.16\textwidth}
    \centering
    \includegraphics[width=\linewidth]{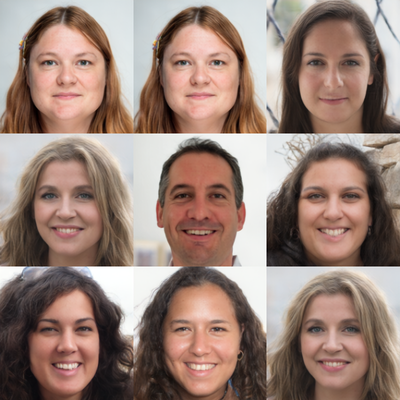}
  \end{subfigure}
  \begin{subfigure}[t]{0.16\textwidth}
    \centering
    \includegraphics[width=\linewidth]{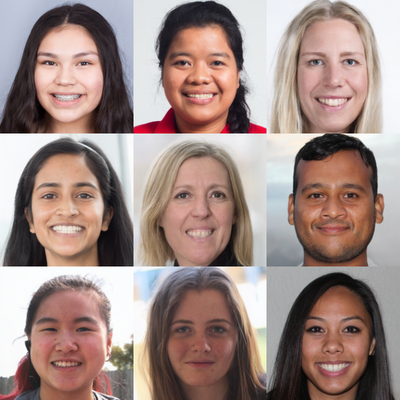}
  \end{subfigure}

  \caption{\textbf{Qualitative comparison between our method and baselines.} While analyzing the images, look for the sharpness of each image and diversity in the content of all images for a method.}
  \label{fig-random-samples-supp}
\end{figure*}

In Figure \ref{fig-random-samples-supp}, we compare the random samples of our method to that of the baseline for more datasets. 

\subsection{Visual Recall}
\begin{figure*}
  \captionsetup[subfigure]{labelformat=parens}
  
  \begin{subfigure}[t]{0.138\textwidth}
    \centering
    Query
  \end{subfigure}
  \begin{subfigure}[t]{0.138\textwidth}
    \centering
    \textit{Ours}
  \end{subfigure}
  \begin{subfigure}[t]{0.138\textwidth}
    \centering
    Ada-IMLE
  \end{subfigure}
  \begin{subfigure}[t]{0.138\textwidth}
    \centering
    FastGAN
  \end{subfigure}
  \begin{subfigure}[t]{0.138\textwidth}
    \centering
    FakeCLR
  \end{subfigure}
  \begin{subfigure}[t]{0.138\textwidth}
    \centering
    FreGAN
  \end{subfigure}
  \begin{subfigure}[t]{0.138\textwidth}
    \centering
    REGAN
  \end{subfigure}
  
  \begin{subfigure}[t]{0.138\textwidth}
    \centering
    \includegraphics[width=\linewidth]{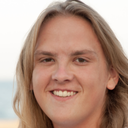}
    \caption{FFHQ-100}
    \label{fig-visual-recall-supp:ffhq1}
  \end{subfigure}
  \begin{subfigure}[t]{0.138\textwidth}
    \centering
    \includegraphics[width=\linewidth]{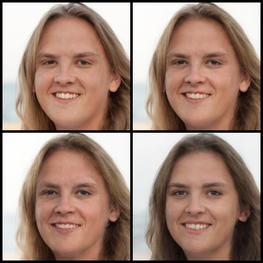}
  \end{subfigure}
  \begin{subfigure}[t]{0.138\textwidth}
    \centering
    \includegraphics[width=\linewidth]{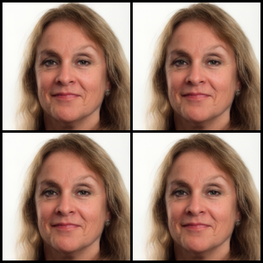}
  \end{subfigure}
  \begin{subfigure}[t]{0.138\textwidth}
    \centering
    \includegraphics[width=\linewidth]{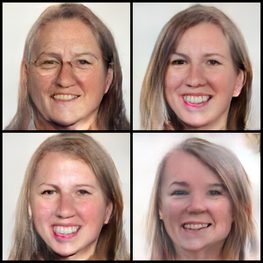}
  \end{subfigure}
  \begin{subfigure}[t]{0.138\textwidth}
    \centering
    \includegraphics[width=\linewidth]{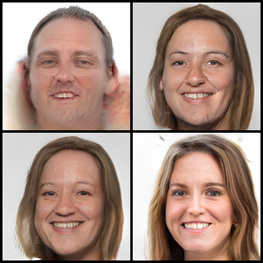}
  \end{subfigure}
  \begin{subfigure}[t]{0.138\textwidth}
    \centering
    \includegraphics[width=\linewidth]{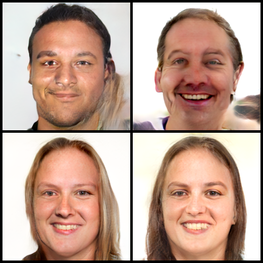}
  \end{subfigure}
  \begin{subfigure}[t]{0.138\textwidth}
    \centering
    \includegraphics[width=\linewidth]{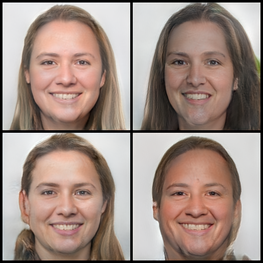}
  \end{subfigure}

  \begin{subfigure}[t]{0.138\textwidth}
    \centering
    \includegraphics[width=\linewidth]{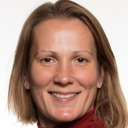}
    \caption{FFHQ-100}
    \label{fig-visual-recall-supp:ffhq2}
  \end{subfigure}
  \begin{subfigure}[t]{0.138\textwidth}
    \centering
    \includegraphics[width=\linewidth]{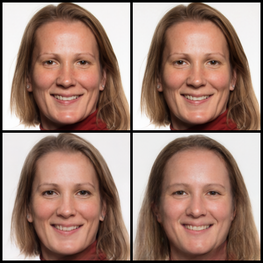}
  \end{subfigure}
  \begin{subfigure}[t]{0.138\textwidth}
    \centering
    \includegraphics[width=\linewidth]{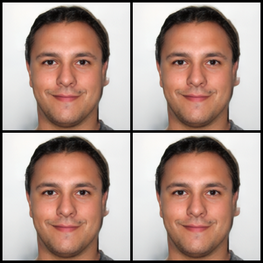}
  \end{subfigure}
  \begin{subfigure}[t]{0.138\textwidth}
    \centering
    \includegraphics[width=\linewidth]{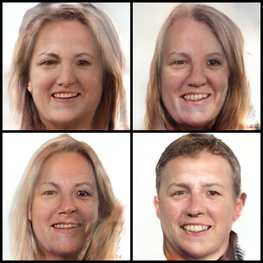}
  \end{subfigure}
  \begin{subfigure}[t]{0.138\textwidth}
    \centering
    \includegraphics[width=\linewidth]{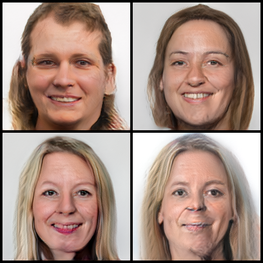}
  \end{subfigure}
  \begin{subfigure}[t]{0.138\textwidth}
    \centering
    \includegraphics[width=\linewidth]{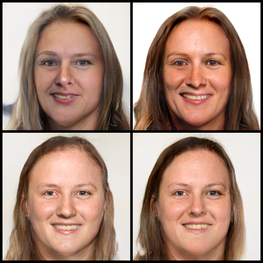}
  \end{subfigure}
  \begin{subfigure}[t]{0.138\textwidth}
    \centering
    \includegraphics[width=\linewidth]{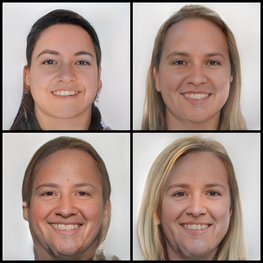}
  \end{subfigure}

  \begin{subfigure}[t]{0.138\textwidth}
    \centering
    \includegraphics[width=\linewidth]{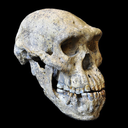}
    \caption{Skulls}
    \label{fig-visual-recall-supp:skull}
  \end{subfigure}
  \begin{subfigure}[t]{0.138\textwidth}
    \centering
    \includegraphics[width=\linewidth]{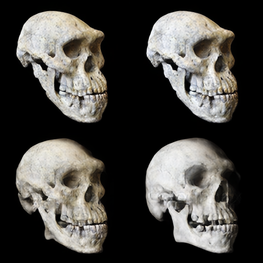}
  \end{subfigure}
  \begin{subfigure}[t]{0.138\textwidth}
    \centering
    \includegraphics[width=\linewidth]{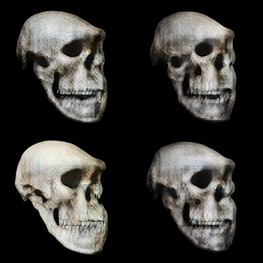}
  \end{subfigure}
  \begin{subfigure}[t]{0.138\textwidth}
    \centering
    \includegraphics[width=\linewidth]{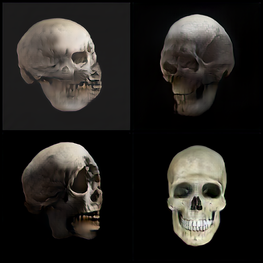}
  \end{subfigure}
  \begin{subfigure}[t]{0.138\textwidth}
    \centering
    \includegraphics[width=\linewidth]{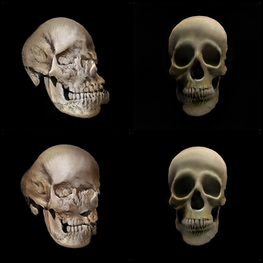}
  \end{subfigure}
  \begin{subfigure}[t]{0.138\textwidth}
    \centering
    \includegraphics[width=\linewidth]{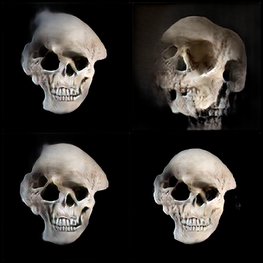}
  \end{subfigure}
  \begin{subfigure}[t]{0.138\textwidth}
    \centering
    \includegraphics[width=\linewidth]{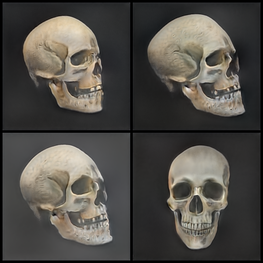}
  \end{subfigure}

  \begin{subfigure}[t]{0.138\textwidth}
    \centering
    \includegraphics[width=\linewidth]{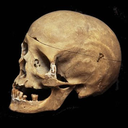}
    \caption{Skulls}
    \label{fig-visual-recall-supp:skull2}
  \end{subfigure}
  \begin{subfigure}[t]{0.138\textwidth}
    \centering
    \includegraphics[width=\linewidth]{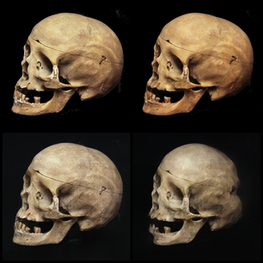}
  \end{subfigure}
  \begin{subfigure}[t]{0.138\textwidth}
    \centering
    \includegraphics[width=\linewidth]{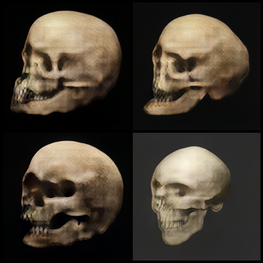}
  \end{subfigure}
  \begin{subfigure}[t]{0.138\textwidth}
    \centering
    \includegraphics[width=\linewidth]{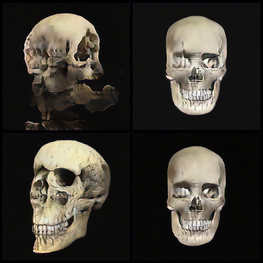}
  \end{subfigure}
  \begin{subfigure}[t]{0.138\textwidth}
    \centering
    \includegraphics[width=\linewidth]{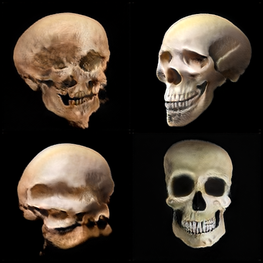}
  \end{subfigure}
  \begin{subfigure}[t]{0.138\textwidth}
    \centering
    \includegraphics[width=\linewidth]{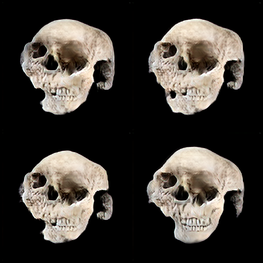}
  \end{subfigure}
  \begin{subfigure}[t]{0.138\textwidth}
    \centering
    \includegraphics[width=\linewidth]{illustrations/queries/query-4/nn_images_ada.png}
  \end{subfigure}

  \begin{subfigure}[t]{0.138\textwidth}
    \centering
    \includegraphics[width=\linewidth]{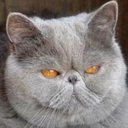}
    \caption{Cat}
    \label{fig-visual-recall-supp:cat}
  \end{subfigure}
  \begin{subfigure}[t]{0.138\textwidth}
    \centering
    \includegraphics[width=\linewidth]{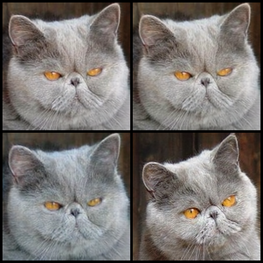}
  \end{subfigure}
  \begin{subfigure}[t]{0.138\textwidth}
    \centering
    \includegraphics[width=\linewidth]{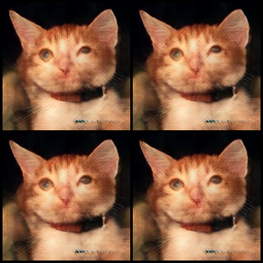}
  \end{subfigure}
  \begin{subfigure}[t]{0.138\textwidth}
    \centering
    \includegraphics[width=\linewidth]{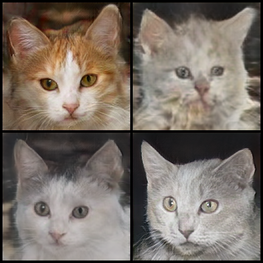}
  \end{subfigure}
  \begin{subfigure}[t]{0.138\textwidth}
    \centering
    \includegraphics[width=\linewidth]{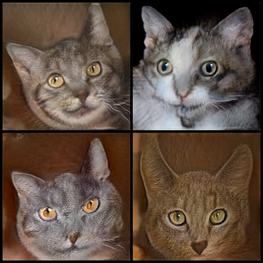}
  \end{subfigure}
  \begin{subfigure}[t]{0.138\textwidth}
    \centering
    \includegraphics[width=\linewidth]{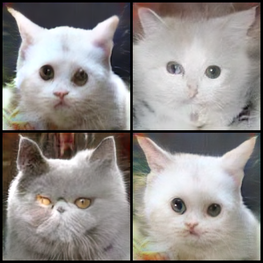}
  \end{subfigure}
  \begin{subfigure}[t]{0.138\textwidth}
    \centering
    \includegraphics[width=\linewidth]{illustrations/queries/query-13/nn_images_ada.png}
  \end{subfigure}

  \begin{subfigure}[t]{0.138\textwidth}
    \centering
    \includegraphics[width=\linewidth]{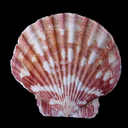}
    \caption{Shells}
    \label{fig-visual-recall-supp:shell}
  \end{subfigure}
  \begin{subfigure}[t]{0.138\textwidth}
    \centering
    \includegraphics[width=\linewidth]{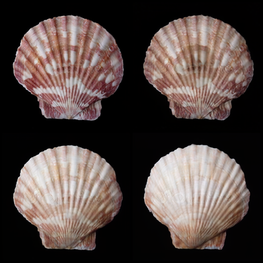}
  \end{subfigure}
  \begin{subfigure}[t]{0.138\textwidth}
    \centering
    \includegraphics[width=\linewidth]{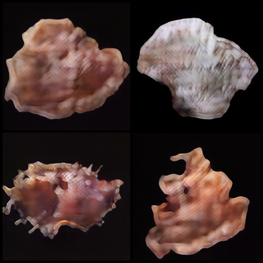}
  \end{subfigure}
  \begin{subfigure}[t]{0.138\textwidth}
    \centering
    \includegraphics[width=\linewidth]{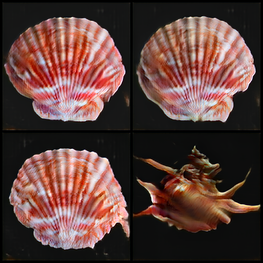}
  \end{subfigure}
  \begin{subfigure}[t]{0.138\textwidth}
    \centering
    \includegraphics[width=\linewidth]{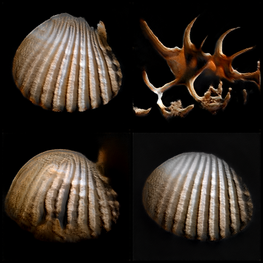}
  \end{subfigure}
  \begin{subfigure}[t]{0.138\textwidth}
    \centering
    \includegraphics[width=\linewidth]{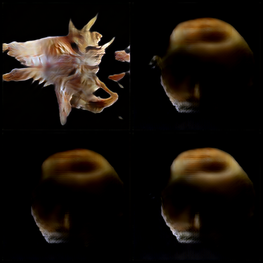}
  \end{subfigure}
  \begin{subfigure}[t]{0.138\textwidth}
    \centering
    \includegraphics[width=\linewidth]{illustrations/queries/query-6/nn_images_ada.png}
  \end{subfigure}

  \begin{subfigure}[t]{0.138\textwidth}
    \centering
    \includegraphics[width=\linewidth]{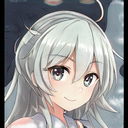}
    \caption{Anime}
    \label{fig-visual-recall-supp:anime}
  \end{subfigure}
  \begin{subfigure}[t]{0.138\textwidth}
    \centering
    \includegraphics[width=\linewidth]{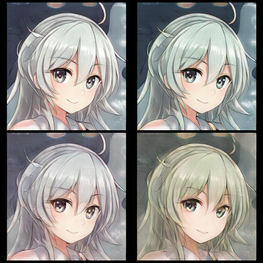}
  \end{subfigure}
  \begin{subfigure}[t]{0.138\textwidth}
    \centering
    \includegraphics[width=\linewidth]{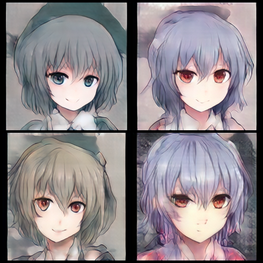}
  \end{subfigure}
  \begin{subfigure}[t]{0.138\textwidth}
    \centering
    \includegraphics[width=\linewidth]{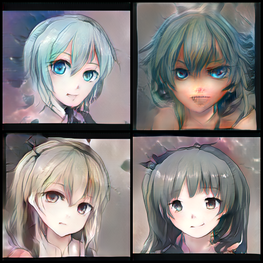}
  \end{subfigure}
  \begin{subfigure}[t]{0.138\textwidth}
    \centering
    \includegraphics[width=\linewidth]{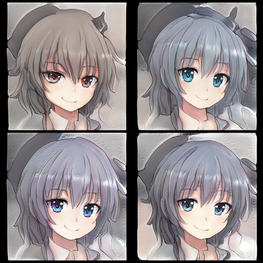}
  \end{subfigure}
  \begin{subfigure}[t]{0.138\textwidth}
    \centering
    \includegraphics[width=\linewidth]{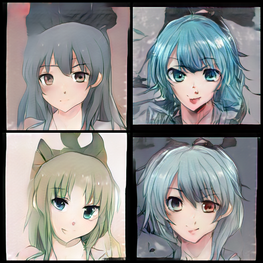}
  \end{subfigure}
  \begin{subfigure}[t]{0.138\textwidth}
    \centering
    \includegraphics[width=\linewidth]{illustrations/queries/query-8/nn_images_ada.png}
  \end{subfigure}

  \begin{subfigure}[t]{0.138\textwidth}
    \centering
    \includegraphics[width=\linewidth]{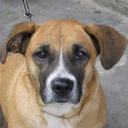}
    \caption{Dog}
    \label{fig-visual-recall-supp:dog}
  \end{subfigure}
  \begin{subfigure}[t]{0.138\textwidth}
    \centering
    \includegraphics[width=\linewidth]{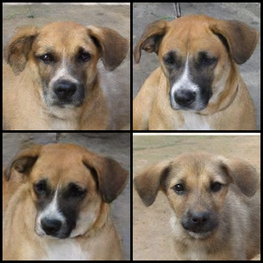}
  \end{subfigure}
  \begin{subfigure}[t]{0.138\textwidth}
    \centering
    \includegraphics[width=\linewidth]{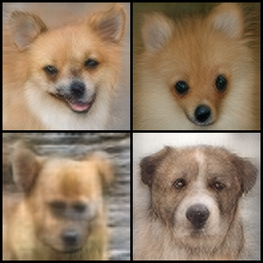}
  \end{subfigure}
  \begin{subfigure}[t]{0.138\textwidth}
    \centering
    \includegraphics[width=\linewidth]{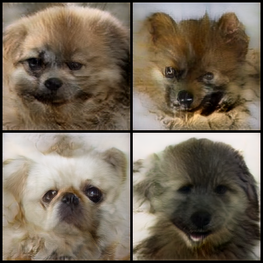}
  \end{subfigure}
  \begin{subfigure}[t]{0.138\textwidth}
    \centering
    \includegraphics[width=\linewidth]{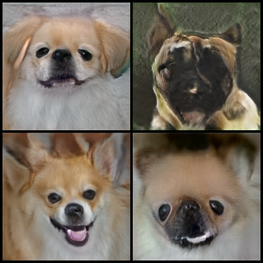}
  \end{subfigure}
  \begin{subfigure}[t]{0.138\textwidth}
    \centering
    \includegraphics[width=\linewidth]{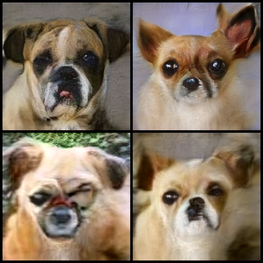}
  \end{subfigure}
  \begin{subfigure}[t]{0.138\textwidth}
    \centering
    \includegraphics[width=\linewidth]{illustrations/queries/query-9/nn_images_ada.png}
  \end{subfigure}

  \caption{\textbf{Visual Recall Test:} First column is the query image from the dataset. Subsequent columns are the samples produced by different methods that are closest to the query image in LPIPS feature space. The samples produced by our method are closer to the query images compared to the baselines, while being sufficiently diverse.}
  \label{fig-visual-recall-supp}
\end{figure*}

Figure \ref{fig-visual-recall-supp} shows the results for the proposed Visual Recall test for more queries. Note how the images produced by our method are the closest to the query and yet have diverse \emph{meaningful} changes. 

Since the images displayed are the \emph{nearest neighbours} of the query images, it would be valuable to emphasize the subtle distinctions in the samples produced by our method. In Figure \ref{fig-visual-recall-supp:ffhq1} and \ref{fig-visual-recall-supp:ffhq2}, we can notice a change in the texture and color of the skin and hair of our samples. In Figure \ref{fig-visual-recall-supp:skull} and \ref{fig-visual-recall-supp:skull2}, we can observe subtle changes to the jaw structure, number of teeth and hue of the different skull samples. Similarly in Figure \ref{fig-visual-recall-supp:cat}, we can notice subtle changes in the color of the fur and tilt of the head for different cat samples. In Figure \ref{fig-visual-recall-supp:anime}, we observe diversity in hair color, background and ear of the produce samples.

\subsection{Ablation on latent dimensions and model parameters}

\begin{table}[h]
\small
\centering
\begin{tabularx}{0.7\linewidth}{nnnnnn}
\toprule
Method  & Dim. & Params. & Anime & Shells & Skulls \\ \midrule
FastGAN & 256      & 29M    & 69.8  & 120.9  & 109.6  \\ \hline
FakeCLR & 512      & 24M    & 77.7  & 148.4  & 106.5  \\ \hline
FreGAN  & 256      & 147M   & 59.8  & 169.3  & 163.3  \\ \hline
ReGAN   & 512      & 24M    & 110.8 & 236.1  & 130.7  \\ \hline
AdaIMLE & 1024     & 36M    & 65.8  & 108.5  & 81.9   \\ \hline
RS-IMLE & 1024     & 36M    & 35.8  & 55.4   & 51.1   \\
        & 512      & 19M    & 48.5 & 52.9   & 60.1   \\
        & 256      & 12M    & 53.8 & 71.7   & 64.3   \\ \bottomrule
\end{tabularx}%
\caption{Comparison between different methods: latent dimensions and number of trainable parameters. Last three columns are FID on Anime, Shells and Skulls dataset.}
\label{tab:arch-table}
\end{table}

Table \ref{tab:arch-table} gives the details about the architectures used by the different methods. 
To decouple the impact of our proposed method (RS-IMLE) from architectural choices, we train using our method using lower latent dimensions. At lower dimensions, the number of parameters for RS-IMLE are significantly lower compared to the other methods. We tabulate the FID for the three most challenging datasets in the last three columns of Table \ref{tab:arch-table}. As we decrease the number of dimensions (and consequently the number of parameters), we observe a slight drop in the FID for our method. However, even at \emph{significantly} lower parameter count, our method \emph{outperforms} the baselines.

\end{document}